\documentclass{article}

\usepackage{arxiv}

\usepackage[utf8]{inputenc} 
\usepackage[T1]{fontenc}    
\usepackage{hyperref}       
\usepackage{url}            
\usepackage{booktabs}       
\usepackage{amsfonts}       
\usepackage{nicefrac}       
\usepackage{microtype}      
\usepackage{graphicx}
\usepackage[authoryear]{natbib}
\usepackage{doi}
\usepackage{pgfplots}
\usepackage{pgf-pie}
\pgfplotsset{compat=newest}
\usepackage{subcaption}
\usepackage{multirow}

\usepackage{subcaption}

\usepackage{amsmath}
\usepackage{enumitem}
\setcitestyle{round}   
\makeatletter
\renewcommand\@makefnmark{\hbox{\normalfont\@thefnmark}}
\renewcommand\@makefntext[1]{\noindent\hbox{\normalfont\@thefnmark}\, #1}
\makeatother
\usepackage{float}
\usepackage{cleveref}

\title{SegCol Challenge: Semantic Segmentation for Tools and Fold Edges in Colonoscopy data}

\author{
\href{https://orcid.org/0000-0002-1137-9357}{\includegraphics[scale=0.06]{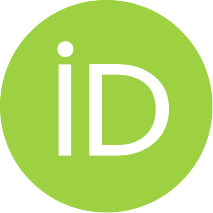}\hspace{1mm}\textnormal{Xinwei Ju}}\footnotemark[1]\hspace{2mm}$^{1,2}$, 
\href{https://orcid.org/0000-0002-0740-7490}{\includegraphics[scale=0.06]{Figs/orcid.pdf}\hspace{1mm}{\textnormal{Rema Daher}}}\footnotemark[1]\hspace{2mm}$^{1,2}$, 
\href{https://orcid.org/0000-0003-0918-0776}{\includegraphics[scale=0.06]{Figs/orcid.pdf}\hspace{1mm}{\textnormal{Razvan Caramalau}}}\footnotemark[1]\hspace{2mm}$^{1,2,3}$,
\href{https://orcid.org/0000-0002-4421-652X}{\includegraphics[scale=0.06]{Figs/orcid.pdf}\hspace{1mm}{\textnormal{Baoru Huang}}}\hspace{1mm}$^{1,2,4}$, \\
\href{https://orcid.org/0000-0002-0908-8972}{\includegraphics[scale=0.06]{Figs/orcid.pdf}\hspace{1mm}{\textnormal{Negin Ghamsarian}}}\hspace{1mm}$^{5}$,
{\hspace{1mm}{\textnormal{Shunsuke Kikuchi}}}\hspace{1mm}$^{6}$,
\href{https://orcid.org/0000-0002-6365-7638}{\includegraphics[scale=0.06]{Figs/orcid.pdf}\hspace{1mm}{\textnormal{Atsushi Kouno}}}\hspace{1mm}$^{6}$,
{\hspace{1mm}{\textnormal{Hiroki Matsuzaki}}}\hspace{1mm}$^{6}$, \\
\href{https://orcid.org/0000-0002-0980-3227}{\includegraphics[scale=0.06]{Figs/orcid.pdf}\hspace{1mm}{\textnormal{Danail Stoyanov}}}\hspace{1mm}$^{1,2}$,
\href{https://orcid.org/0000-0002-4609-1177}{\includegraphics[scale=0.06]{Figs/orcid.pdf}\hspace{1mm}{\textnormal{Francisco Vasconcelos}}}\hspace{1mm}$^{1,2}$
}

\date{
\small 
$^{1}$UCL Hawkes Institute, University College London, London, UK\\
$^{2}$Department of Computer Science, University College London, London, UK\\
$^{3}$Medtronic, London, UK\\
$^{4}$Department of Computer Science, University Liverpool, Liverpool, UK\\ 
$^{5}$ARTORG Center for Biomedical Engineering Research, University of Bern, Switzerland\\ 
$^{6}$Jmees Inc., Kashiwa, Chiba, JP\\ 
\texttt{\{xinwei.ju.22, rema.daher.20\}@ucl.ac.uk}
}

\hypersetup{
    colorlinks=true,            
    linkcolor=black,            
    urlcolor=black,            
    citecolor=blue,             
    pdfborder={0 0 0},         
    pdftitle={SegCol Challenge: Semantic Segmentation for Tools and Fold Edges in Colonoscopy Data},
    pdfauthor={Xinwei Ju, Rema Daher, et al.},
    pdfsubject={Colonoscopy Semantic Segmentation},
    pdfkeywords={Colonoscopy, Semantic Segmentation, Tools, Fold Edges}
}

\begin{document}
\maketitle
\footnotetext[1]{These authors contributed equally to this work and are co-first authors.}


\begin{abstract}
Improving the reliability and completeness of colonoscopic inspection is critical for reducing missed lesions and improving colorectal cancer prevention. Reliable scene understanding in colonoscopy is essential for improving navigation, reconstruction, and assessment of inspection completeness. In particular, anatomical structures such as mucosal folds provide stable geometric cues that may support endoscope localization, while surgical instruments introduce dynamic occlusions that complicate visual interpretation. However, existing gastrointestinal endoscopy datasets largely focus on disease detection or artifact segmentation, leaving a gap in precise annotations of structural landmarks and instruments in real colonoscopic imagery.

In this work, we introduce \textbf{SegCol}, a new dataset and benchmark for semantic segmentation of colon fold edges and surgical instruments derived from the EndoMapper dataset. The dataset contains manually annotated pixel-level masks for three instrument classes and thin fold-edge structures across temporally consistent image sequences. SegCol forms the basis of the \textit{SegCol Challenge} organized as part of the EndoVis Challenge at MICCAI 2024, which evaluates both supervised segmentation performance and annotation-efficient active learning strategies. Details are available at \url{https://www.synapse.org/Synapse:syn54124209/wiki/626563}, and project page with code at \url{https://github.com/surgical-vision/segcol_challenge}.

We conduct a systematic study of segmentation metrics, analyzing how commonly used measures such as Dice, ODS/OIS, AP, and CLDice respond to different structural perturbations and object geometries. In addition, we provide an evaluation of participating methods and analyze architectural design choices and active learning strategies. Our findings highlight that metric behavior strongly depends on structural properties of the target, suggesting that evaluation protocols should be carefully selected when assessing segmentation performance in endoscopic scenes.

\end{abstract}


\section{Introduction}

Colorectal cancer (CRC) still remains the second leading cause of cancer-related deaths worldwide~\citep{bray2024global}. Colonoscopy is the gold standard screening method for preventing CRC, involving an endoscopic visual inspection of the colon and detection of polyps that may develop into cancer~\citep{lee2017risk}. However, polyp detections can be missed if the endoscopic inspections are not thorough enough~\citep{ahn2012miss}. These misses often arise from incomplete visualization of the mucosal surface due to complex colon geometry, variable illumination, camera motion, and occlusions from folds or surgical instruments~\citep{van2019role,brand2017missed,rees2016expert}. Consequently, improving the reliability and completeness of colon inspection remains a critical challenge in gastrointestinal endoscopy.

To address these limitations, significant research efforts have been devoted to developing navigation systems for colonoscopy that localise the endoscopic camera within the colon anatomy, build a map with detected polyp locations, and quantify colon inspection completeness~\citep{schmidt2024tracking}. While significant progress has been achieved through the adaptation of Simultaneous Localisation and Mapping (SLAM) pipelines to colonoscopy~\citep{ma2021rnnslam,morlana2024topological}, they are still not reliable enough for clinical translation, with significant failure modes related to varying image quality, dynamic occlusion, deforming tissue, repetitive textures, specular reflections, and low-texture regions. Navigation systems, therefore, require support from further visual scene understanding in order to evaluate which image regions can be utilised as reliable navigation landmarks. Semantic segmentation in colonoscopy has mostly focused on polyps~\citep{gupta2024systematic,manan2025advancement,liu2025polypsensed}, since this has direct diagnostic value. However, comprehensive navigation of the colon requires the recognition of more frequent anatomical landmarks and the presence of instruments~\citep{kim2024density,taghiakbari2023automated,rueckert2024methods}.

In this paper we report on the methodology, data, results, and findings of the SegCol EndoVis Challenge presented at MICCAI 2024. This challenge focused on the semantic segmentation of anatomy landmarks and instruments during colonoscopy video examinations. These structures have significant impact on the visual tracking of landmarks for endoscopic navigation and their reliable segmentation enables future potential to develop semantic SLAM navigation algorithms with increased robustness in colonoscopy. In terms of anatomical landmarks, we focus on the detection of colon fold edges, which have been investigated as stable landmarks for depth perception and local navigation. We also investigate the semantic segmentation of endoscopic instruments used during screening and polyp removal (forceps, balloon dilator, and captivator), which may occlude and modify local anatomy landmarks~\citep{kurniawan2017flexible}. We aim at evaluating the state-of-the-art for this multi-label semantic segmentation problem, with especial focus on assessing the sensitivity of algorithms and evaluation metrics. We observe that due to the presence of widely different semantic shapes, from large areas (balloon), to extremely thin structures (colon edges, captivator snares) both the algorithms and evaluation metrics require careful assessment. 

We introduce a novel set of manually annotated pixel-level semantic labels for a subset of the EndoMapper dataset \citep{Azagra_2023}, which contains videos of complete colonoscopies. We labelled a total of 8,440 images, sampled as batches of 40 consecutive frames in the range 10-12 FPS, with pixel-level masks of 3 instrument classes and 1 class representing the edges of colon folds. The labels have been made available through incorporation into the publicly available EndoMapper Dataset (\url{https://www.synapse.org/Synapse:syn26707219/wiki/615178}. While large amounts of colonoscopy video data are available, the manual labelling of fine semantic structures is extremely labour intensive, and therefore we also investigate automated strategies for efficient sampling of training data through active learning techniques, with the aim of establishing best practices for future data labelling in this domain.

By releasing the SegCol dataset, challenge benchmark, and evaluation framework, we aim to facilitate future research on structural scene understanding in colonoscopy and to support the development of navigation systems capable of more reliable endoscopic localization and mapping. Our contributions can be summarised as following:
\begin{itemize}[itemsep=0.2pt, topsep=1pt]
    \item We introduce the \textbf{SegCol dataset}, a new benchmark for colonoscopy scene understanding with pixel-level annotations of colon fold edges and multiple surgical instrument classes on 8,440 images from the EndoMapper dataset.
    \item We present the methodology of the \textbf{SegCol Challenge} from MICCAI 2024, including two tasks: 1) supervised multi-label semantic segmentation of anatomy and instruments in colonoscopy; 2) active learning strategies for efficiently sampling training data.
    \item We provide a \textbf{comprehensive evaluation and analysis} of participating methods, examining design choices. We observe that utilising separate heads for instruments and anatomical structures generally provides more robust results.
    \item We conduct a \textbf{systematic study of segmentation metrics}, revealing how commonly used metrics for both semantic segmentation and semantic edge detection (Dice, ODS/OIS, AP, and CLDice) respond differently to structural errors such as dilation, spatial misalignment, and topology disruption. The sensitivity of different metrics is highly correlated with the defining shape characteristics of the different instruments and colon folds.
\end{itemize}

\section{Related Work}

\subsection{Surgical Instrument Segmentation}



Surgical instrument segmentation is a widely studied problem, with many available public benchmarks and frequent updates to its state-of-the-art methods. Attention mechanisms are frequently utilised to focus on the most relevant features while suppressing irrelevant information and noise \citep{wang2021surgical,ghamsarian2021recal,yang2022attention,shen2023branch}. Other relevant methodological trends include leveraging multi-scale information \citep{ceron2022real,mahmood2022dsrd} and instrument motion priors \citep{jin2019incorporating,qin2019surgical}. Common segmentation architecture backbones include DeepLabV3, DeepPyramid+ \citep{ghamsarian2024deeppyramid+}, UNet++ \citep{zhou2018unet} and its variants like U-NetPlus \citep{hasan_linte_2019}. Recently, many works have shifted toward Unsupervised Domain Adaptation (UDA) to address the scarcity of ground-truth annotations in surgical scenes \citep{li2025unsupervised,colleoni2021robotic,wang2023generalizing,luo2023adaptnet}. Most surgical tool evaluation protocols rely on well established metrics robust to class imbalances, such as the Dice score. However, ~\citep{karimi2021lossodyssey} highlights the limitations of Dice in capturing the alignment of thin structures or fine contour details. Therefore, boundary-aware metrics such as Chamfer distance or topology-aware metrics such as clDice (~\citep{shit2021cldice} provide alternative means to evaluate the segmentation of shapes with specific characteristics. 

The segmentation of instruments in colonoscopy is significantly less explored than in other medical intervention contexts, and public benchmarks for this domain are scarce. The EndoCV Challenge \citep{ali2021deep} released a multi-label segmentation dataset which includes multiple artifact classes (specularities, blood, blur) and instruments as a single class. The dataset includes around 800 images with semantic segmentation labels, but only a fraction includes instruments. Kvazir-Instrument \citep{jha2021kvasir} provides 590 frames labelled with instrument segmentation masks as a single class. This data was later incorporated into the MedAI 2021 Challenge for polyp and instrument segmentation as part of a wider multi-class dataset \citep{Jha2021Validating}. In contrast, our dataset provides 3,120 labelled frames containing instruments, grouped into 3 different classes: forceps, captivator (polyp snare), and balloon. While the banchmarking in EndoCV and MedAI rely on metrics related to pixel-level statists (Dice and mIoU), we observe that instrument classes such as the captivator include very thin structures and therefore warrant the investigation of topology-aware metrics such as clDice~\citep{shit2021cldice} for this problem.


\subsection{Colon Anatomy Segmentation}

Semantic segmentation of anatomical structures in colonoscopy overwhelmingly focuses on polyps due to its direct application to diagnosis \citep{borgli2020hyperkvasir, ali2021deep, vazquez2017benchmark, bernal2017comparative}. Beyond conventional binary polyp segmentation, NeoPolyp \citep{nguyen2021neounet} further reformulates the task as a fine-grained benchmark for joint polyp segmentation and neoplasm detection by distinguishing neoplastic and non-neoplastic polyps. However, the segmentation of more frequently observed colon landmarks for the purpose of continuous navigation is comparatively less studied. Lumen segmentation \citep{vazquez2017benchmark} can provide partial information about endoscopic view orientation. However, this does not represent a fixed landmark and its boundaries are not always well defined. FoldIt~\citep{mathew2021foldit} proposes manual labelling and supervised segmentation of colon haustral folds. These are well defined landmarks within the colon that naturally create visibility occlusions and may hide polyps behind them. They proposes dense labelling of fold surface regions between their outer edges and the colon wall. While outer edges are visually well defined, their connection to the colon wall may be ambiguous on 2D images due to lighting and perspective. This leads to inconsistent segmentation boundaries and temporally unstable inference in continuous video. Subsequent work by \citep{jin2023self} has shown that segmenting thin fold edges rather than dense regions produces temporally more stable detections. However, this approach is still limited by the lack of large-scale labelled edge data and mostly relies on self-supervision and indirect evaluation techniques. C3VD \citep{bobrow2023colonoscopy} is a phantom colonoscopy dataset with accurate groundtruth of colon 3D shape and endoscope trajectories. This dataset contains occlusion segmentation masks generated from geometric projection of the colon shape onto the endoscope view. While these occlusion masks are highly correlated with haustral fold locations, they are not temporally stable landmarks, as their shape changes with viewpoint. Furthermore, there is a noticeable domain gap between phantom C3VD data and real colonoscopy, which affects model generalisability. This motivates us to further pursue this direction and manually label a large volume of temporally consistent pixel-thin fold edges on real colonoscopy clips in order to evaluate fold edge detection more comprehensively.

\subsection{Edge Detection}

Edges have already been shown to enhance downstream tasks in endoscopy. Scene edges have been used as auxiliary cues for polyp segmentation \citep{bui2024meganet}, and \citep{ju2026multimodalmonocularendoscopicdepth} incorporate edge detection to improve pose and depth estimation. These examples highlight that edges are meaningful structural signals that can strongly support navigation, reconstruction, and other geometry-driven tasks.


Edge detection is commonly evaluated using the BSDS protocol~\citep{martin2004bsds}, which summarizes precision–recall behavior via ODS, OIS, and AP. ODS reports the best dataset-wide threshold, OIS selects the best per-image threshold, and AP measures threshold-free ranking quality~\citep{arbelaez2011gpb}. Deep edge detectors such as HED~\citep{xie2015hed} and RCF~\citep{liu2017rcf} standardize these metrics, which are also adopted in medical boundary analysis where small contour deviations can be clinically meaningful. However, their differential sensitivity to boundary-specific errors remains insufficiently examined. 


\subsection{Active Learning}
With the increasing data requirements of deep learning models, active learning has gained significant attention as a strategy to reduce annotation effort through efficient selection of training data. Fundamentally, active learning approaches rely either on exploiting model uncertainty \citep{uncsamp, Kirsch2019BatchBALD:Learning} or on ensuring sample diversity \citep{Sener2017ActiveApproach, near2}. More recently, these two principles have been combined through learning-based acquisition models specifically designed for sample selection \citep{Yoo2019LearningLearning, ugcn, kim2021task}.

Colonoscopy invariably requires careful training data selection from long exploratory videos containing an abundance of redundant and uninformative frames. Despite this, the application of active learning in this domain remains under-explored. One of the early applications of active learning in colonoscopy was proposed for automatic polyp detection by \citep{polyp_zhe_guo}. Their uncertainty-based sampling strategy focuses on incorporating false positive detections as negative training examples. \citep{taal_endo} extended the active learning method of \citep{kim2021task} to polyp segmentation and depth estimation tasks. Their sampling strategies were evaluated on the Kvasir-SEG dataset \citep{jha2021kvasir} and the Colonoscopy Depth dataset \citep{anita_synth}. However, these datasets are already a well curated selection of labelled images, and therefore the challenge of sampling data from colonoscopy sequences with large amounts of redundancy and low quality frames remains unaddressed.

\section{Dataset details}

Our dataset was first released as part of the SegCol EndoVis challenge at MICCAI 2024 (excluding test data labels). The full dataset, including hidden labels, is now integrated with the EndoMapper dataset repository (\url{https://www.synapse.org/Synapse:syn26707219}).

\subsection{Dataset Sampling} 

All images are sampled from EndoMapper dataset videos as batches of 40 consecutive frames at 10-12 FPS.  The sampling of consecutive frames proved fundamental to label and review fold edges consistently across illumination changes. While most frames are sampled during endoscope withdrawal (imaged in air), we also include frames from endoscope insertion (imaged in fluid). We sampled around 150 batches (6000 images) where instruments are present, and 150 batches (6000 images) without any instrument. 

Batches containing instruments were sampled using the EndoMapper's labels endoscopist comments, in the form of timestamped text transcripts, which include notes of instrument usage. Batches without instruments are sampled to guarantee that most of its frames are clear and well-lit. We utilise an endoscopic image quality classifier \citep{barbed2022semantic} to identify "clear" frames and sample batches around them. Each batch will contain at least a few good quality frames, however, the remaining may present challenges (defocus, occlusion, poor illumination, etc). Finally, a manual review of all sampled batches was performed to curate the final dataset.

\begin{figure}[ht]
    \centering
    \includegraphics[width=\linewidth]{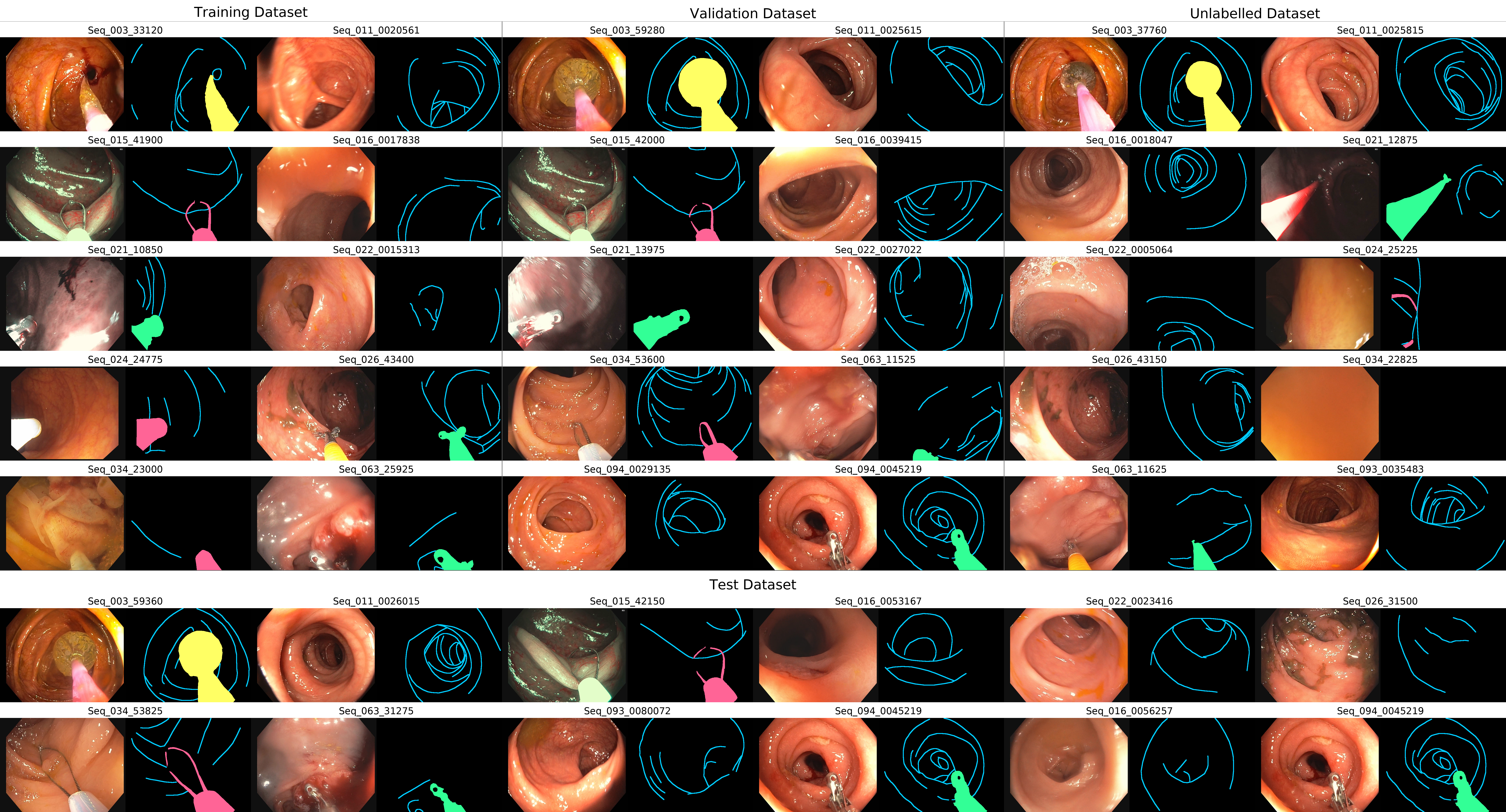}
    \caption{\textbf{Grid visualization of dataset samples.} The top five rows display representative samples from the Training, Validation, and Unlabeled sets, organized into two columns per category to highlight data diversity. The bottom two rows provide a comparative view of the Test dataset, maintaining consistent formatting to allow for visual inspection of domain alignment across all splits.}
    \label{fig:datsetgrid}
\end{figure}

\subsection{Dataset annotation}

Images are labeled with pixel-level masks of 3 instrument classes (captivator, balloon, forceps) and 1 class representing the edges of colon folds. 

Fold edges are annotated as 1-pixel thickness contours, precisely delineating natural landmarks where there’s a scene depth discontinuity in the colon anatomy. Edges were first annotated as polyline vertices and then converted to pixel masks. Visibility challenges include occlusion by bubbles, reflections, water, instruments, and very dark regions. Additional annotation challenges are caused when the exact start-end points of edges are ambiguous. Ambiguity was addressed by preserving temporal consistency across 40-frame batches. Extreme cases of inconsistency or ambiguity were not labeled (or deleted during review). 

Surgical tools are annotated with dense pixel masks covering diverse instruments, including resection devices, forceps, and balloons, with special attention to intricate shapes, such as thin wires, holes, and tissue occlusions. Visibility challenges include saturated reflections and a small visible area (\textit{e.g.} tip of the instrument in image corner). 

\begin{figure}[ht]
    \centering
    \includegraphics[width=\linewidth]{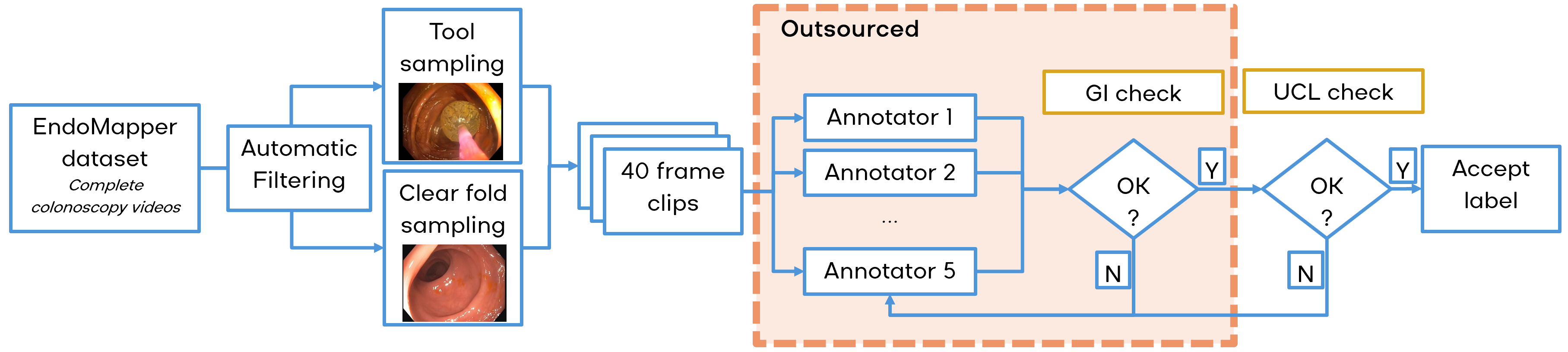}
    \caption{The Annotation Pipeline.}
    \label{fig:annotation_pipeline}
\end{figure}

The complete annotation pipeline is shown in Figure~\ref{fig:annotation_pipeline}. The labelling involved an external team of 5 annotators over 3 months. During the first month, the team included a Gastroenterologist to review the labelling workflow. Each batch of 40 frames was assigned to a single annotator. Each labelled batch was reviewed by the authors by either accepting it or raising issues for label correction and subsequent iterative review. All images were labelled at 1440 x 1080 resolution and then down-sampled to 640x480 for release. 

\subsection{SegCol Challenge}

Within the context of the SegCol Endovis Challenge (\url{https://www.synapse.org/Synapse:syn54124209/wiki/626563}), the dataset is organised in the following manner:
\begin{itemize} 
\item A labelled training set of 2,000 images
\item  A labelled validation set of 560 images
\item An unlabelled training dataset of 5,320 images 
(labels withheld during the challenge).
\item A test set of 560 images (data not released during the challenge)
\end{itemize}

\begin{figure}[h!]
    \centering
    \includegraphics[width=\linewidth]{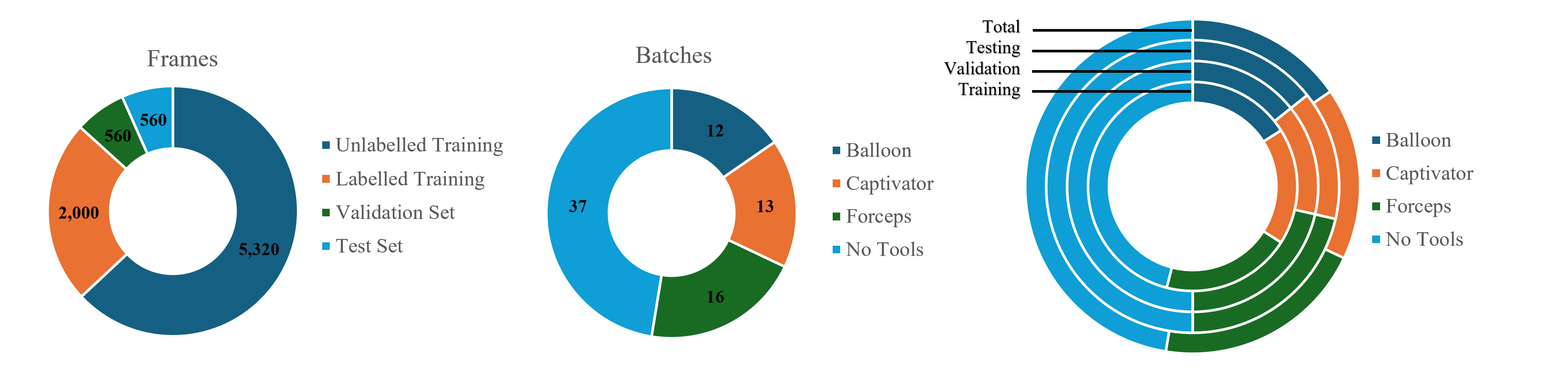}
    \caption{\textbf{Dataset Composition and Tool Distribution.} This figure provides an overview of the dataset structure via three ring charts. The left chart illustrates the proportional split of the unlabeled training, labeled training, validation, and test sets. The center chart displays the distribution of 40-frame batches categorized by tool type (Balloon, Captivator, Forceps) or background frames, which include instances with tissue folds or plain tissue views without surgical instruments. The right chart highlights the distribution of these tools across the different data subsets to demonstrate class balance and ensure experimental consistency.}
    \label{fig:dataset_ring_charts}
\end{figure}

 The total number of labelled samples are 78 batches, divided into 50 for training, 14 for validation, and 14 for testing. The distribution of tools across different sets are shown in Table~\ref{table:tool_distribution}. This breakdown provides an evenly distributed dataset for the three tool categories, facilitating comprehensive model training and evaluation. Dataset composition and tool distribution is shown in Figure~\ref{fig:dataset_ring_charts}. Dataset samples are visualized in Figure~\ref{fig:datsetgrid}

\begin{table}[ht]
\centering
\caption{Distribution of tools across training, validation, and testing sets.}
\begin{tabular}{lcccc}
\toprule
\textbf{Tool Type}       & \textbf{Training Set} & \textbf{Validation Set} & \textbf{Testing Set} & \textbf{Total} \\ \hline
Balloon          & 8                     & 2                       & 2                    & 12             \\ 
Captivator       & 9                     & 2                       & 2                    & 13             \\ 
Forceps          & 10                    & 3                       & 3                    & 16             \\ 
Total Batches & 50                    & 14                      & 14                   & 78             \\ \bottomrule
\end{tabular}

\label{table:tool_distribution}
\end{table}

\begin{table}[ht]
\centering
\caption{Summary of the full SegCol dataset released in the EndoVis SegCol Challenge.}
\begin{tabular}{lccc}
\toprule
\textbf{Dataset Split} & \textbf{\# Frames}& \textbf{Labels During Challenge?} & \textbf{Labels Publicly Released Now?}\\
\midrule
Training Set (Labelled) & 2,000& Yes& Yes
\\
Validation Set& 560& Yes& Yes
\\
Training Set (UnLabeled) & 5,320 & (withheld)& Yes
\\
Test Set& 560& (withheld)& Yes
\\
\midrule
Total & 8,440 & -- & Yes \\
\bottomrule
\end{tabular}
\label{tab:segcol_full_dataset}
\end{table}

The SegCol Challenge comprised two tasks:
\begin{itemize}
    \item Task 1 - Multi-class Segmentation: Participants design segmentation architectures using the labeled training data, with a focus on accurately and robustly segmenting anatomical features and surgical instruments.
    \item Task 2 - Active Learning: Participants develop sampling strategies to select the most informative 400 frames from the unlabeled training set, addressing the practical challenges of efficient dataset labeling.
\end{itemize}

In order to test a diverse range of evaluation metrics, participants were asked to submit semantic segmentation models that output confidence logits rather than thresholded segmentations.

\section{Evaluation}
\label{sec:Eval}
\begin{figure}
    \centering
    \includegraphics[width=\linewidth]{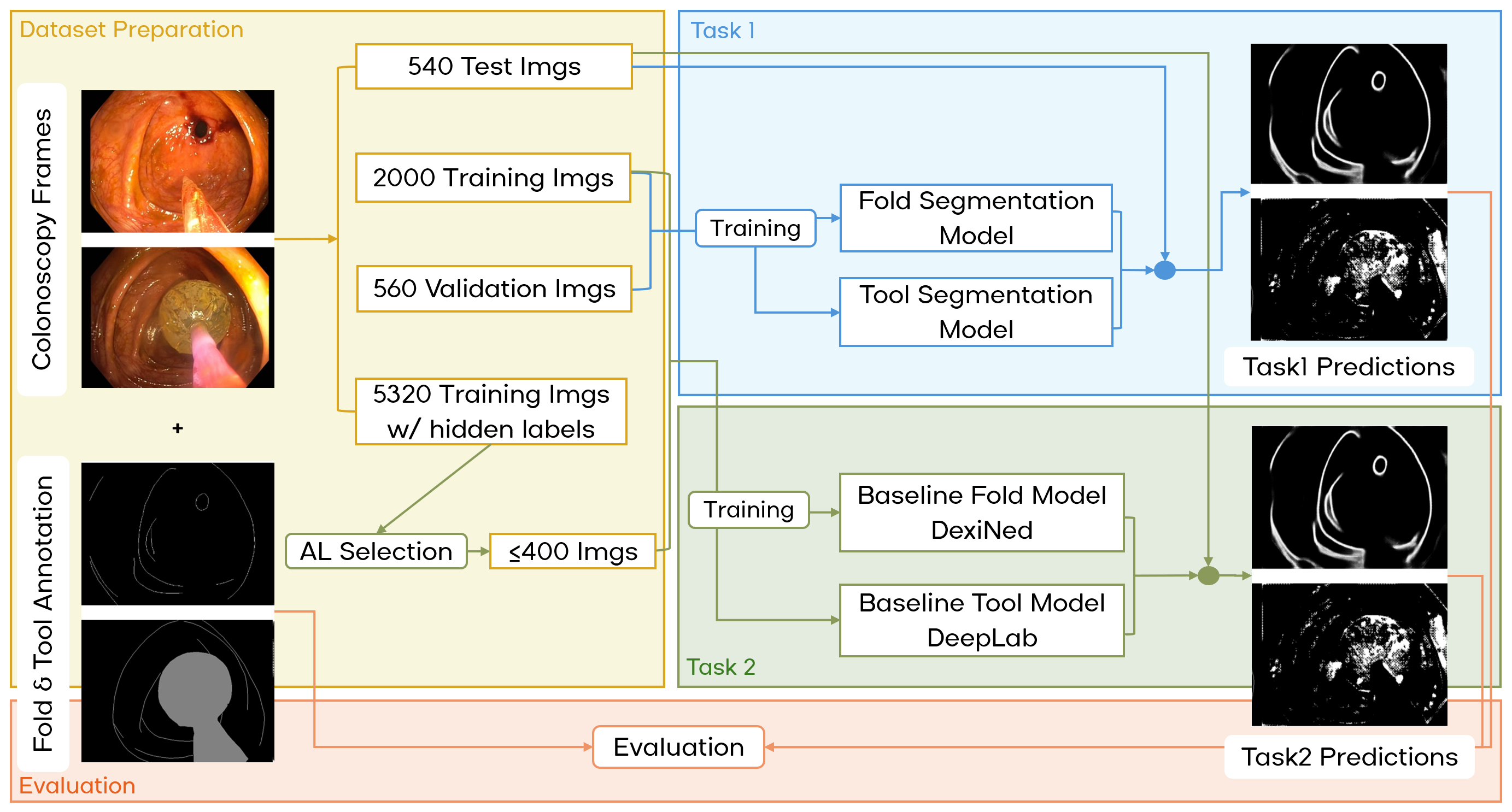}
    \caption{\textbf{SegCol Evaluation Workflow.} The flowchart places data preparation on the left block, with Task 1 and Task 2 represented by the upper and lower blocks on the right.} 
    \label{fig:eval_flow}
\end{figure}

Our evaluation workflow is shown in \ref{fig:eval_flow}. Semantic segmentation is most commonly evaluated through pixel-level metrics (e. g. Dice). However, the presence of very diverse target shapes, including both very thin (colon folds, captivator snare) and thick structures (balloon, forceps, captivator shaft) warrants a more thorough investigation of evaluation metrics for this problem. We consider metrics for standard semantic segmentation (Dice), semantic segmentation of thin structures (clDice), edge detection (ODS, OIS), and object detection (AP). 

\begin{enumerate}

\item \textbf{Dice Coefficient} measures the overlap between predicted and ground truth segmentations:
\begin{equation}
    \mathrm{Dice} = \frac{2|X \cap Y|}{|X| + |Y|} 
    = \frac{2TP}{2TP + FP + FN}
\end{equation}
where $X$ and $Y$ denote the predicted and ground truth segmentations, respectively, and $TP$, $FP$, $FN$ represent true positives, false positives, and false negatives.

\textbf{Global Dice (micro over pixels)} aggregates statistics over all images before computing the score:
\begin{equation}
\mathrm{Dice}_{\mathrm{micro}} 
= 
\frac{2 \sum_{i=1}^{N} TP_i}
{2 \sum_{i=1}^{N} TP_i + \sum_{i=1}^{N} FP_i + \sum_{i=1}^{N} FN_i},
\end{equation}
where $i$ indexes images (or frames) and $N$ is the total number of images. 
This formulation weights images proportionally to their foreground pixel counts.

\item \textbf{Average Precision (AP)} evaluates performance across varying thresholds by integrating the precision–recall curve:
\begin{equation}
\mathrm{Precision} = \frac{TP}{TP + FP}, 
\quad
\mathrm{Recall} = \frac{TP}{TP + FN},
\end{equation}
\begin{equation}
AP = \sum_{n} (R_n - R_{n-1}) P_n,
\end{equation}
where $P_n$ and $R_n$ denote precision and recall at the $n$-th threshold.

\item \textbf{Optimal Image Scale (OIS) and Optimal Dataset Scale (ODS)} evaluate performance under threshold variation using the F1 score:
\begin{equation}
F1 = \frac{2 \cdot \mathrm{Precision} \cdot \mathrm{Recall}}
    {\mathrm{Precision} + \mathrm{Recall}}, 
F1(i,t)
=
\frac{2TP_{i,t}}{2TP_{i,t} + FP_{i,t} + FN_{i,t}},
\end{equation}
\begin{equation}
OIS = \frac{1}{N} \sum_{i=1}^{N} \max_{t} F1(i,t),
\end{equation}
\begin{equation}
ODS = \max_{t} \frac{1}{N} \sum_{i=1}^{N} F1(i,t),
\end{equation}
where $i$ indexes images and $t$ denotes the threshold.

\item \textbf{Centerline Dice (CLDice)} specifically addresses the evaluation of thin structures using skeleton consistency:
\begin{equation}
\mathrm{CLDice}
=
\frac{2 \cdot TPR_{cl} \cdot PPV_{cl}}
{TPR_{cl} + PPV_{cl}},
\end{equation}
where $TPR_{cl}$ and $PPV_{cl}$ are computed from skeletonized masks.

\end{enumerate}

We also consider the following implementation details:

\begin{enumerate}

\item \textbf{Zero-handling:} Special consideration is given to cases where both prediction and ground truth contain no positive pixels, assigning perfect scores to avoid undefined metrics. For each class, such cases were identified using thresholds determined by Youden's Index on the ROC curve:
\begin{equation}
    J = \max_t (TPR(t) - FPR(t))
\end{equation}

\item \textbf{Multi-class Evaluation:} The framework supports both binary (fold vs. tool) and four-class evaluation (fold and three tool types). For binary evaluation, tool classes are merged through maximum operation:
\begin{equation}
    Tool_{binary} = \max(Tool_1, Tool_2, Tool_3)
\end{equation}

\item \textbf{Optimal Threshold Selection:} Inherently, OIS and ODS rely on a set of thresholds (100 thresholds spanning from 0.01 to 0.99). According to the ODS results and the relevant threshold, we determine the optimal threshold, which is then applied to other metrics (i.e., Dice, CLDice, and AP).

\item \textbf{Participant ranking:} Challenge participants were ranked based on mean Average Precision (mAP) across classes, computed as:
\begin{equation}
    mAP = \frac{1}{C}\sum_{c=1}^C AP_c
\end{equation}
where C is the number of classes. This choice was justified by mAP being a well established and documented metric in evaluating both pixel-thin edges and thicker objects. 

In this paper we further provide an empirical assessment of evaluation metrics on this dataset.

\item \textbf{Absence-Aware Evaluation} Two approaches exist for aggregating segmentation metrics across a dataset: pixel pooling, where all pixels from all images are combined into a single TP/FP/FN count before computing the score, and per-frame averaging, where the metric is computed independently per image and then averaged. To evaluate a model's ability to correctly predict the absence of a tool in a frame, a clinically relevant capability, zero handling is required: frames where both the ground truth and prediction are empty should be scored as perfect rather than undefined. This zero handling is well-suited to per-frame averaging, where each frame contributes a score of 1.0 to the mean without distorting the overall distribution. For pixel pooling, however, the same handling injects large blocks of artificial true-positive pixels that overwhelm the genuine signal and skew scores toward 1. For this reason, we report both pooled metrics, Dice and Precision, which follow common practice and weight frames proportionally to object size, and per-frame metrics, ODS, OIS, and CLDice, which treat each frame equally and correctly reward empty-frame predictions through explicit zero handling.

\end{enumerate}


\section{Methods}
\begin{figure}
    \centering
    \includegraphics[width=\linewidth]{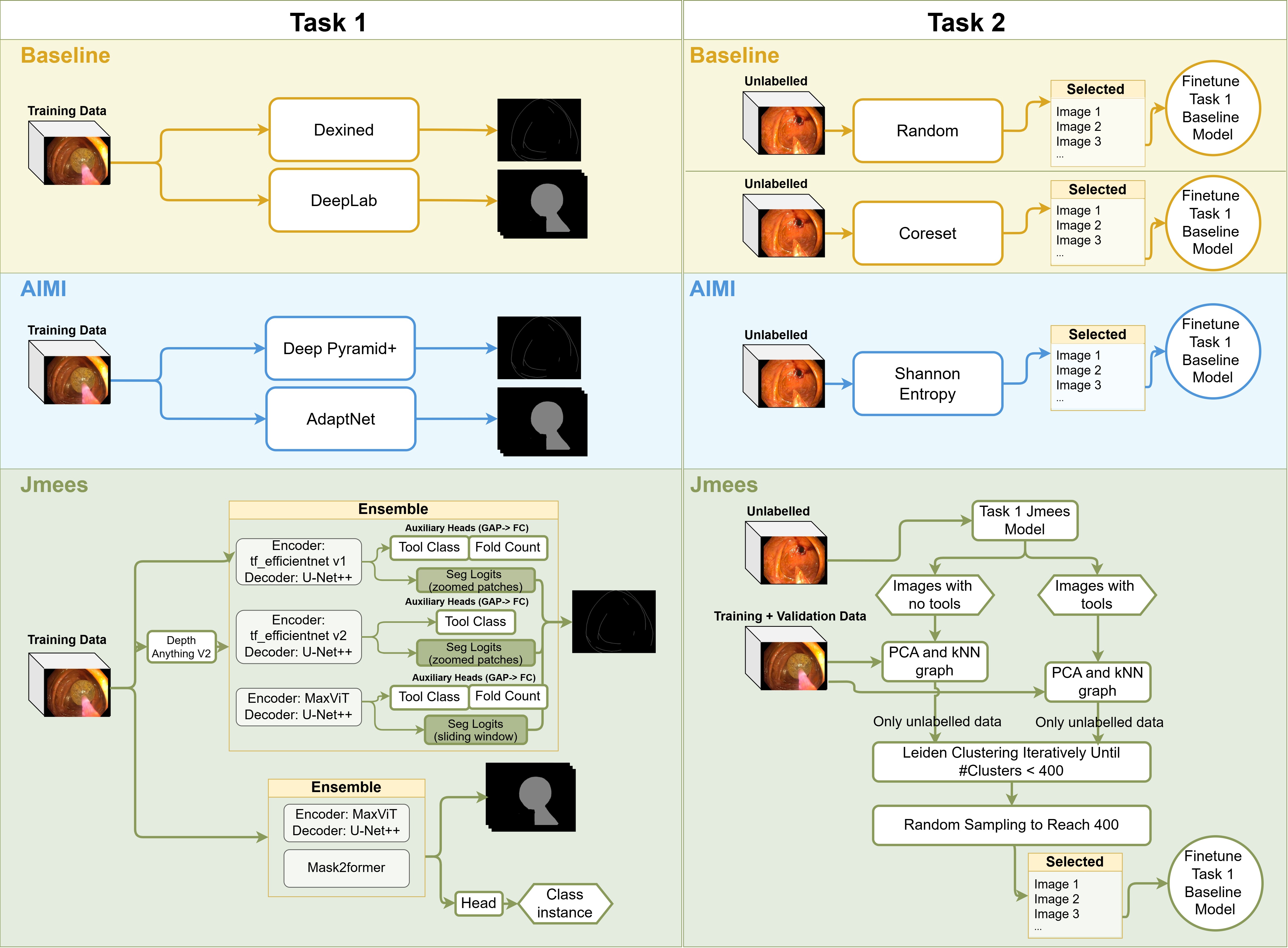}
    \caption{Flowcharts of the baseline and participants' methods for Task 1 and 2.}
    \label{fig:placeholder}
\end{figure}

\subsection{Task 1: Multi-class Segmentation}
\subsubsection{Baseline}
Task 1 addresses multiclass segmentation of colonoscopy images with four target classes: folds, captivator, forceps, and balloon. Because anatomical folds appear primarily as thin, elongated boundaries while surgical instruments form dense image regions, we adopt a hybrid approach that combines edge detection and semantic segmentation. Fold segmentation is formulated as an edge detection task, while instrument segmentation is treated as standard pixel-wise semantic segmentation.

For fold detection, we use DexiNed \citep{soria2023dense}, an edge detection network trained directly on the task’s training data using default architecture and hyperparameter settings. Ground-truth fold annotations are used to supervise the prediction of fold edges. For instrument segmentation, we train a DeepLab~\citep{chen2017deeplab} model in a multiclass setting to predict dense masks for the captivator, forceps, and balloon classes. The two models are trained independently, each optimized for its respective target structures, with no shared parameters or joint training.

At inference time, each input frame is processed by both networks. DexiNed produces a fold-edge probability map, while DeepLab outputs per-pixel class predictions for the instrument classes. The resulting predictions are merged in a post-processing step to form a single multiclass segmentation output, in which fold edges and instrument regions are jointly represented. This approach leverages specialized representations for different anatomical and tool structures while producing a unified output compatible with the Task 1 formulation.

\subsubsection{Team AIMI}

\textbf{Compared Architectures.}
Team AIMI evaluates four encoder--decoder architectures that have demonstrated effectiveness in surgical video analysis. 
DeepLabV3+~\citep{chen2018encoder} is used as a strong baseline segmentation model with atrous spatial pyramid pooling and a ResNet-50 encoder. 
ReCalNet~\citep{ghamsarian2021recal} is a region--channel joint calibration network designed for challenging medical image conditions, including blurred boundaries, strong reflections, transparency, and motion blur. 
Its ReCal module refines feature representations by modeling intra-region dependencies for local structural coherence, inter-region dependencies for global spatial relationships, and channel--region cross dependencies between feature channels and spatial contexts. 
It also enforces semantic consistency by aligning features corresponding to the same anatomical label across different views. 
AdaptNet~\citep{ghamsarian2021lensid} is a U-Net--based encoder--decoder framework composed of Cascade Pooling Fusion (CPF), Shape/Scale-Adaptive Feature Fusion (SSF), and Feature Fusion Decision (FFD) modules. 
CPF extracts multi-scale features using cascaded pooling, SSF uses separate branches to adapt to object size and shape variations, and FFD applies pixel-level attention to control the contribution of each feature map. 
DeepPyramid+~\citep{ghamsarian2024deeppyramid+} uses Pyramid View Fusion (PVF) to mimic human visual perception and enhance pixel-level information, together with Deformable Pyramid Reception (DPR), which employs dilated deformable convolutions for shape- and scale-adaptive feature extraction.

\textbf{Data Processing and Augmentation.}
All input images are resized to $512\times512$ pixels. 
Weak augmentations consist of random rescaling between $0.5$ and $2.0$ of the original size and horizontal flipping with probability $0.5$. 
Strong augmentations further apply color jittering with brightness/contrast factor $0.6$ and saturation factor $0.4$, random grayscale conversion with probability $0.2$, and a random choice of Gaussian blur or sharpness adjustment.

\textbf{Experimental Setup.}
For both instrument and fold segmentation, AIMI instantiates the architectures with different encoders, including \textit{VGG-16}, \textit{ResNet-34}, and \textit{ResNet-50}. 
All encoders are initialized with ImageNet-pretrained weights. 
Training is performed using stochastic gradient descent (SGD). 
The randomly initialized segmentation heads are optimized with an initial learning rate $\text{lr}_{\text{init}}\in\{0.005, 0.002, 0.01, 0.02\}$, while the encoder learning rate is set to $0.1\,\text{lr}_{\text{init}}$. 
A polynomial learning rate decay with power $0.9$ is applied over $T$ iterations:
\begin{equation}
    \text{lr}_t = \text{lr}_{\text{init}} \left(1 - \frac{t}{T}\right)^{0.9}.
\end{equation}
All models are trained for $80$ epochs. 
Performance is evaluated using Dice coefficient and Intersection-over-Union (IoU), and the best learning rate per architecture--backbone combination is reported.

\textbf{Ablation Study.}
The team conducts ablation studies over four architectures and multiple backbone choices for both fold and instrument segmentation. 
The results are listed in Table~\ref{tab:aimi_ablation}. 

\begin{table*}[ht]
\centering
\caption{Ablation study of state-of-the-art methods on fold and instrument segmentation from AIMI (Dice / IoU, \%).
Bold values indicate the best-performing method--backbone combination for Fold and Tools-Avg, respectively (row-wise comparison).}
\label{tab:aimi_ablation}
\footnotesize
\begin{tabular*}{0.85\linewidth}{@{\extracolsep{\fill}} l l c c c c c c c c}
\toprule
\textbf{Task} & \textbf{Backbone} 
& \multicolumn{2}{c}{\textbf{DeepLabV3+}} 
& \multicolumn{2}{c}{\textbf{ReCal-Net}}
& \multicolumn{2}{c}{\textbf{AdaptNet}} 
& \multicolumn{2}{c}{\textbf{DeepPyramid+}} \\
& 
& Dice & IoU & Dice & IoU & Dice & IoU & Dice & IoU \\
\midrule
\multirow{3}{*}{\textbf{Fold}}
& VGG16    
  & --   & --   & 26.81 & 16.56 & 29.34 & 18.43 & 28.98 & 18.78 \\
& ResNet34 
  & --   & --   & 28.47 & 18.01 & 29.37 & 18.87 & \textbf{29.64} & \textbf{19.11} \\
& ResNet50 
  & 25.23 & 16.32 & --   & --   & --   & --   & --   & --   \\
\midrule
\multirow{3}{*}{\textbf{Tools-Avg}}
& VGG16    
  & -- & -- 
  & 80.61 & 69.29 
  & 90.19 & 82.73 
  & 87.17 & 78.16 \\
& ResNet34 
  & -- & -- 
  & 86.40 & 76.81 
  & \textbf{90.40} & \textbf{82.95} 
  & 79.43 & 68.34 \\
& ResNet50 
  & 84.72 & 74.65
  & -- & --
  & -- & -- 
  & 89.87 & 82.40 \\
\midrule\midrule
\multirow{3}{*}{\textbf{Captivator}}
& VGG16    
  & --   & --   & 82.47 & 70.17 & 89.73 & 81.37 & 87.52 & 77.81 \\
& ResNet34 
  & --   & --   & 85.49 & 74.66 & 89.89 & 81.64 & 77.69 & 63.52 \\
& ResNet50 
  & 81.38 & 68.60 & --   & --   & --   & --   & 90.83 & 83.20 \\
\midrule
\multirow{3}{*}{\textbf{Forceps}}
& VGG16    
  & --   & --   & 64.65 & 47.76 & 82.80 & 70.65 & 77.11 & 62.75 \\
& ResNet34 
  & --   & --   & 77.89 & 63.79 & 83.93 & 72.31 & 62.76 & 45.73 \\
& ResNet50 
  & 75.45 & 60.57 & --   & --   & --   & --   & 80.46 & 67.30 \\
\midrule
\multirow{3}{*}{\textbf{Balloon}}
& VGG16    
  & --   & --   & 94.70 & 89.94 & 98.04 & 96.16 & 96.87 & 93.92 \\
& ResNet34 
  & --   & --   & 95.82 & 91.98 & 97.38 & 94.89 & 97.84 & 95.77 \\
& ResNet50 
  & 97.32 & 94.77 & --   & --   & --   & --   & 98.32 & 96.71 \\
\bottomrule
\end{tabular*}
\end{table*}

\textbf{Architecture Selection and Final Submission.}
The ablation study shows that the optimal architecture differs between anatomical fold segmentation and instrument segmentation. 

For fold segmentation, DeepPyramid+ with a ResNet-34 encoder achieves the best Dice and IoU scores, reaching $29.64\%$ Dice and $19.11\%$ IoU. 
AdaptNet is close on fold segmentation, especially with ResNet-34 ($29.37\%$ Dice and $18.87\%$ IoU), but DeepPyramid+ remains slightly stronger. 
This is consistent with the design of DeepPyramid+, where Pyramid View Fusion enhances pixel-level representations and Deformable Pyramid Reception uses dilated deformable convolutions to adapt to the shape and scale variations of thin fold structures.

For tool segmentation, AdaptNet with ResNet-34 obtains the highest average tool performance, with $90.40\%$ Dice and $82.95\%$ IoU. 
It also performs consistently well across the individual instrument classes: $89.89\%$ Dice for captivator, $83.93\%$ Dice for forceps, and $97.38\%$ Dice for balloon. 
This supports the use of its CPF, SSF, and FFD modules, which respectively provide multi-scale feature extraction, shape/scale-adaptive fusion, and pixel-level attention-based feature selection. 
DeepPyramid+ with ResNet-50 performs strongly for some individual tool classes, achieving $90.83\%$ Dice for captivator and $98.32\%$ Dice for balloon, but its average tool performance remains slightly lower than AdaptNet with ResNet-34. 
ReCalNet improves over simpler configurations in some settings but does not achieve the best performance in either fold or average tool segmentation, suggesting that region--channel calibration alone is less effective than explicit pyramid or shape/scale-adaptive fusion for this task.

Based on these ablations, AIMI adopts a task-specific final submission strategy: DeepPyramid+ with ResNet-34 is used for fold segmentation, while AdaptNet with ResNet-34 is used for instrument segmentation.

\subsubsection{Team Jmees}
Jmees splits the task into two subtasks: instrument segmentation plus classification, and fold segmentation. 
Both subtasks are implemented using encoder--decoder segmentation frameworks with task-specific architectural adaptations, together with auxiliary supervision and boundary-aware optimisation strategies.

\textbf{Fold segmentation.}
For fold segmentation, a multi-scale ensemble composed of three segmentation models is adopted, all based on the \textit{U-Net++} architecture with different encoder backbones. 
All models take as input the RGB image concatenated with depth estimated by \textit{DepthAnythingV2}~\citep{yang2024depth}. 

Model~1 and Model~2 share the same encoder--decoder structure with \textit{EfficientNet}~\citep{tan2019efficientnet} and \textit{EfficientNetV2}~\citep{tan2021efficientnetv2} encoders, respectively, paired with a U-Net++ decoder. 
They are trained using different resolution sets: $72, 104, 176, 480, 1376$ for Model~1 and $32, 64, 96, 224, 640$ for Model~2. 
This multi-resolution strategy exposes the models to diverse scale contexts, improving sensitivity to extremely thin fold boundaries.

Model~3 follows a MaxViT-base encoder~\citep{tu2022maxvit} with a U-Net++ decoder~\citep{zhou2018unet,zhou2019unet} and performs sliding-window inference on ratio-preserving padded images. 
Fold predictions are computed on zoomed patches and aggregated to form the final output. 
Additionally, a padded mask is applied to the output logits to constrain predictions to the valid endoscopic field of view, reducing peripheral false positives. The configurations of the three fold segmentation models are summarised in Table~\ref{tab:jmees_fold_model_table}.

During inference, the three fold models output probability maps which are then averaged:
\begin{equation}
P_{\mathrm{final}}
=
\frac{1}{3}(P_1 + P_2 + P_3)
\quad .
\end{equation}

This ensemble combines multi-resolution global context from Models~1--2 with patch-level zoomed detail from Model~3, improving the detection of thin anatomical fold structures.

\textbf{Tool segmentation.}
For tool segmentation, two complementary models specialised for multi-class instrument recognition are employed. 
The first model combines a \textit{MaxViT}~\citep{tu2022maxvit} encoder together with a \textit{U-Net++}~\citep{zhou2018unet,zhou2019unet} decoder. 
Input images are resized to $512 \times 512$, and a masking strategy is applied to remove the black padded regions outside the valid endoscopic field of view. 
The decoder output is converted into pixel-wise class probabilities using softmax.

The second model adopts a \textit{Mask2Former} architecture, which provides transformer-based instance-aware segmentation and complements the convolutional \textit{MaxViT--U-Net++} model in handling complex spatial layouts. 
Similar to fold segmentation, tool segmentation also adopts an ensemble strategy, combining the MaxViT--U-Net++ output with the finetuned Mask2Former prediction to obtain the final tool segmentation result.

\textbf{Auxiliary supervision.}
Both subtasks employ auxiliary semantic supervision to improve optimisation stability and enhance global semantic consistency.

For fold segmentation, all three models incorporate an auxiliary tool classification head attached to the deepest encoder layer. 
A global average pooling operation is applied to the bottom encoder feature map, producing a global feature vector of dimension $D$, where $D$ denotes the number of encoder output channels. 
The pooled representation is passed through a fully connected layer to produce a multi-label prediction over all tool classes in the dataset. 
Each output dimension corresponds to one tool category and predicts whether that tool is present in the image. 
The head therefore outputs a vector of size $C_{\text{tools}}$.

In addition, Model~1 and Model~3 in fold segmentation include an extra regression head that predicts the number of folds in the image. 
This head takes the same globally pooled encoder representation as input and maps it through a fully connected layer to produce a single scalar output representing the predicted fold count, treated as a continuous value. 
These auxiliary supervision signals provide additional global semantic constraints on the encoder representation, which empirically stabilise optimisation and improve convergence speed.

For tool segmentation, considering that only one instrument appears per frame, an auxiliary classification head is also used to predict the instrument category from the encoder representation, reducing confusion between visually similar tool classes. 

\textbf{Boundary-aware supervision and optimisation.}
For both subtasks, the segmentation objective generally combines Dice, Focal, and boundary-aware losses:
\begin{equation}
\mathcal{L}_{\mathrm{Segmentation}}
=
\lambda_{\mathrm{dice}}\mathcal{L}_{\mathrm{Dice}}
+
\lambda_{\mathrm{focal}}\mathcal{L}_{\mathrm{Focal}}
+
\mathcal{L}_{\mathrm{boundary}}.
\end{equation}

The Dice and Focal components are given by
\begin{equation}
\mathcal{L}_{\mathrm{Dice}}
=
1 -
\frac{2\sum PG+\epsilon}
{\sum P+\sum G+\epsilon},
\quad
\mathcal{L}_{\mathrm{Focal}}
=
-\alpha(1-p_t)^\gamma \log(p_t),
\end{equation}
where $P$ is the predicted probability map, $G$ is the ground-truth mask, and $\epsilon$ is a small constant for numerical stability. 
Here, $p_t$ denotes the predicted probability of the ground-truth class, $\gamma$ controls the focusing strength, and $\alpha$ balances class contributions. 
The Dice and Focal terms improve region-level overlap and hard-pixel learning.

For fold segmentation, given that fold boundaries are often only 1--2 pixels wide, customised boundary-aware supervision is used to encourage sharper, topologically consistent edge predictions. 
The ground-truth boundary map is converted into a Gaussian-blurred gradient representation. 
Formally, let $B$ be the binary boundary mask, and $\tilde{B} = \mathrm{GaussianBlur}(B)$ the softened boundary target. 
Let $P$ denote the predicted boundary probability. 
A generic form of the fold boundary-aware loss is:
\begin{equation}
\mathcal{L}_{\mathrm{boundary}}
=
1 -
\frac{2 \sum P \tilde{B}}
{\sum P + \sum \tilde{B}}
\quad .
\end{equation}
Certain fold model variants additionally employ HED-style edge supervision for enhanced boundary refinement, as summarised in Table~\ref{tab:jmees_fold_model_table}. 
Model~1 in fold segmentation additionally incorporates a false-positive suppression term to penalise spurious fold predictions.

For tool segmentation, the boundary-aware term follows the same edge-focused motivation, but uses the tool boundary target described in the original method. 
For numerical stability, the stabilisation constant $\epsilon$ is added to the boundary-aware term to avoid division by zero. 
The overall objective further combines the segmentation loss with an auxiliary classification loss:
\begin{equation}
\mathcal{L}_{\text{total}}
=
\mathcal{L}_{\text{Segmentation}}
+
\mathcal{L}_{\text{Classification}}.
\end{equation}

The auxiliary classification head in tool segmentation is supervised using binary cross-entropy with logits over the tool categories:
\begin{equation}
\mathcal{L}_{\mathrm{Classification}}
=
\frac{1}{C}
\sum_{c=1}^{C}
\mathrm{BCEWithLogits}(z_c,y_c),
\end{equation}
where $z_c$ is the predicted logit for class $c$, $y_c$ is the corresponding ground-truth label, and $C$ denotes the number of tool classes.

\begin{table}[t]
\centering
\setlength{\tabcolsep}{1.3pt}
\caption{Comparison of Fold Segmentation Model Configurations and Performance in Jmees Method}
\label{tab:jmees_fold_model_table}
\footnotesize

\begin{tabular*}{\linewidth}{@{\extracolsep{\fill}} lc|c|c|c}
\toprule
\textbf{Model} & \textbf{Backbone + Decoder} & \textbf{Auxiliary Head} & \textbf{Input Modalities} & \textbf{Loss Functions} \\
\midrule
Model~1
& EfficientNet-L2 + U-Net++ 
& Tool classification + Fold count 
& RGB + Depth 
& Dice + Focal + Boundary + False-positive \\
\midrule
Model~2
& EfficientNetV2 + U-Net++ 
& Tool classification 
& RGB + Depth 
& Dice + Focal + HED \\
\midrule
Model~3 
& MaxViT-base + U-Net++ 
& Tool classification + Fold count 
& RGB + Depth + Mask 
& Dice + Focal + HED + Boundary \\
\bottomrule
\end{tabular*}

\vspace{2pt}

\begin{flushleft}
\footnotesize \textit{Note.} mAP for Model~1, Model~2, and Model~3 are 0.3313, 0.3314, and 0.1534, respectively.
\end{flushleft}
\end{table}

\textbf{Ablation study.}
The ablation study for tool segmentation (Table~\ref{tab:jmees_tool_ablation}) shows progressive improvement from adding the auxiliary head, replacing SE-ResNeXt-50 (32$\times$4d)~\citep{hu2018squeeze} with MaxViT, applying valid-region masking, and introducing boundary-aware supervision. 
The final MaxViT--U-Net++ model achieved an overall validation score of 0.991, and ensembling it with the finetuned Mask2Former further improved the score to 0.9986.

\begin{table}[ht]
    \centering
    \caption{Tool Segmentation and Classification Sub-task Ablation Study (Jmees)}
    \label{tab:jmees_tool_ablation}
    \begin{tabular*}{0.94\linewidth}{@{\extracolsep{\fill}} lccccccc}
        \toprule
        & DexiNed & SE-ResNeXt-50 & Remove folds & Aux & MaxViT & \textit{Masking} & $\mathcal{L}_{\mathrm{boundary}}$ \\
        \midrule
        Forceps & 0.133 & 0.522 & 0.406 & 0.828 & 0.9074 & 0.8567 & 0.9868 \\
        Resec & 0.033 & 0.703 & 0.851 & 0.762 & 0.9090 & 0.9579 & 0.9883 \\
        Balloon & 0.652 & 0.682 & 0.988 & 0.964 & 0.9838 & 0.9988 & 0.9988 \\
        Overall & 0.272 & 0.636 & 0.748 & 0.851 & 0.935 & 0.937 & 0.991 \\
        \bottomrule
    \end{tabular*}
\end{table}

\subsection{Task 2: Active Learning for Segmentation}
\subsubsection{Baseline}
We consider two baseline strategies for active learning in the segmentation setting. The first baseline employs \emph{random sampling}, where a subset of unlabeled samples is selected uniformly at random for annotation. The second baseline uses \emph{coreset sampling}, which aims to select a representative and diverse subset of samples by minimizing redundancy in the feature space. The methodology follows \citep{Sener2017ActiveApproach}. For both baselines, the selected samples are annotated and used to fine-tune the baseline segmentation model introduced in Task~1. This fine-tuning process constitutes the active learning step, allowing the model to iteratively improve its performance by leveraging selectively acquired labeled data.

\subsubsection{Team AIMI}

To reduce labeling effort and improve generalization, AIMI integrates their segmentation models into a pseudo-labeling-based active learning framework. 
The key idea is to estimate prediction reliability using entropy and rank unlabeled images by informativeness in an unsupervised manner.

Let $D = \{x_1, x_2, \dots, x_M\}$ denote the pool of unlabeled images. 
Two pretrained segmentation networks are used: 
$f^{\text{inst}}$ for instrument segmentation and $f^{\text{fold}}$ for fold segmentation.

\textbf{Per-pixel entropy.}
For each image $x_i$, two forward passes are computed.

For instrument segmentation (multi-class, $C=4$), a softmax probability map is produced:
\begin{equation}
p_{ij}^{\text{inst},c} = \text{softmax}(f^{\text{inst}}(x_i))^c.
\end{equation}
The corresponding pixel-wise Shannon entropy is
\begin{equation}
S(p_{ij}^{\text{inst}}) =
- \sum_{c=1}^{4}
p_{ij}^{\text{inst},c}
\log p_{ij}^{\text{inst},c}.
\end{equation}

For fold segmentation (binary), the sigmoid output
\begin{equation}
p_{ij}^{\text{fold}} = \sigma(f^{\text{fold}}(x_i))
\end{equation}
is used to compute binary entropy:
\begin{equation}
S(p_{ij}^{\text{fold}}) =
- p_{ij}^{\text{fold}} \log p_{ij}^{\text{fold}}
- (1 - p_{ij}^{\text{fold}}) \log (1 - p_{ij}^{\text{fold}}).
\end{equation}

\textbf{Image-level uncertainty aggregation.}
Let $N_i$ denote the number of pixels in image $x_i$. 
The two entropy maps are fused with a weighting factor of 3 favouring instrument uncertainty:
\begin{equation}
\bar{S}(x_i) =
\frac{1}{4 N_i}
\sum_{j=1}^{N_i}
\left[
3\, S(p_{ij}^{\text{inst}})
+
S(p_{ij}^{\text{fold}})
\right].
\end{equation}

\textbf{Sample selection and pseudo-labeling.}
Images with low $\bar{S}(x_i)$ are considered reliable and may be exploited through pseudo-labeling without additional manual annotation. 
Conversely, images with high $\bar{S}(x_i)$ indicate high predictive uncertainty and are prioritised for manual annotation.

After computing $\bar{S}(x_i)$ for all images, the 400 most uncertain samples are selected by ranking in descending order:
\begin{equation}
S^* =
\left\{
x_i
\mid
i \in
\text{argsort}(\bar{S}, \text{descending})[:400]
\right\}.
\end{equation}

In practice, the framework alternates between 
(i) training on the current labeled pool (augmented with pseudo-labeled low-entropy samples), and 
(ii) querying new high-entropy samples for annotation, iteratively refining the model.

\subsubsection{Team Jmees}

For Task 2, Jmees employ an active-learning pipeline designed to select the most informative and diverse frames from the unlabeled dataset for training. Their strategy consists of a multi-step filtering pipeline.

\begin{enumerate}
    \item  They first reduce the raw collection of 4,000 images to approximately 800 by retaining only frames containing surgical instruments, using the tool-segmentation model developed in Task 1.
    \item The subset was further condensed to roughly 300 images by applying Principal Component Analysis (PCA) with \texttt{n\_components=1000} to project images into a compact embedding that preserves dominant appearance variations while suppressing noise, constructing a $k$-Nearest Neighbors (kNN) graph in this reduced space to capture local similarity structure, and clustering this graph with the Leiden algorithm under a strict maximum community size constraint (\texttt{max\_comm\_size}=2), which enforces very small, diverse communities and thus discourages redundant sampling.
    \item For non–instrument frames, a similar \textit{PCA–kNN–Leiden} procedure is applied to capture variation in background and anatomical context. Unlike the instrument branch, the community-size constraint is gradually relaxed: \texttt{max\_comm\_size} starts at $2$ and increases stepwise (every five iterations in the implementation) until the number of eligible clusters falls below the quota required to complete the 400–frame budget. This controlled relaxation merges overly fine clusters into broader yet coherent groups, improving diversity among non–tool views.
    \item Finally, frames are sampled from the union of the tool-containing and non-tool clusters to assemble a balanced set of 400 images, ensuring that the final annotation subset captures both instrument-related variability and broader anatomical and visual diversity.
\end{enumerate}


\section{Results and Discussion}
This section summarises the participants' results across the evaluation metrics described in \Cref{sec:Eval}. 



\subsection{Task 1} 

\begin{table}[ht]
\centering
\caption{SegCol Task1 Quantitative Results \\ The quantitative results of baseline and participating teams are compared, with the best performances \textbf{bolded} and the second best \underline{underlined}.}
\label{tab:segcol_table1}
\begin{tabular*}{\linewidth}{@{\extracolsep{\fill}} cccccccc}

\toprule
\textbf{Classes} & \textbf{Methods} & \textbf{ODS}$\uparrow$ & \textbf{OIS}$\uparrow$ & \textbf{Dice}$\uparrow$ & \textbf{AP}$\uparrow$ & \textbf{CLDice}$\uparrow$ & \textbf{Threshold} \\
\midrule
\multirow{3}{*}{Folds} 
& Dexined 
& \underline{0.250} & \underline{0.258} 
& \underline{0.265} 
& \underline{0.170} 
& \underline{0.279} 
& 0.830 \\

& AIMI 
& 0.197 & 0.208 
& 0.218 
& 0.105 
& 0.205 
& 0.570 \\

& Jmees 
& \textbf{0.340} & \textbf{0.361} 
& \textbf{0.361} 
& \textbf{0.274} 
& \textbf{0.372} 
& 0.440 \\

\midrule
\multirow{3}{*}{Tools}
& Deeplab 
& 0.235 & 0.281 
& 0.660 
& 0.699 
& 0.245 
& 0.650 \\

& AIMI 
& \underline{0.795} & \underline{0.797} 
& \underline{0.921} 
& \underline{0.887} 
& \underline{0.793} 
& 0.010 \\

& Jmees 
& \textbf{0.924} & \textbf{0.928} 
& \textbf{0.981} 
& \textbf{0.998} 
& \textbf{0.984} 
& 0.410 \\

\midrule\midrule
\multirow{3}{*}{Captivator} 
& Deeplab 
& 0.077 & 0.085 
& 0.507 
& 0.430 
& 0.286 
& 0.670 \\

& AIMI 
& \underline{0.537} & \underline{0.537} 
& \underline{0.772} 
& \underline{0.719} 
& \underline{0.922} 
& 0.010 \\

& Jmees 
& \textbf{0.665} & \textbf{0.667} 
& \textbf{0.959} 
& \textbf{0.984} 
& \textbf{0.988} 
& 0.400 \\

\midrule
\multirow{3}{*}{Forceps} 
& Deeplab 
& 0.058 & 0.082 
& 0.168 
& 0.113 
& 0.253 
& 0.550 \\

& AIMI 
& \underline{0.563} & \underline{0.564} 
& \underline{0.699} 
& \underline{0.591} 
& \underline{0.784} 
& 0.010 \\

& Jmees 
& \textbf{0.610} & \textbf{0.613} 
& \textbf{0.899} 
& \textbf{0.974} 
& \textbf{0.938} 
& 0.010 \\

\midrule
\multirow{3}{*}{Balloon} 
& Deeplab 
& 0.680 & 0.681 
& 0.829 
& 0.923 
& 0.918 
& 0.660 \\

& AIMI 
& \textbf{0.995} & \textbf{0.995} 
& \textbf{0.987} 
& \textbf{0.992} 
& \textbf{0.994} 
& 0.710 \\

& Jmees 
& \underline{0.924} & \underline{0.924} 
& \textbf{0.987} 
& \textbf{1.000} 
& \underline{0.968} 
& 0.420 \\

\toprule
\end{tabular*}
\end{table}

\begin{figure}[ht]
    \centering
    \begin{subfigure}[t]{0.49\textwidth}
        \centering
        \includegraphics[width=\linewidth]{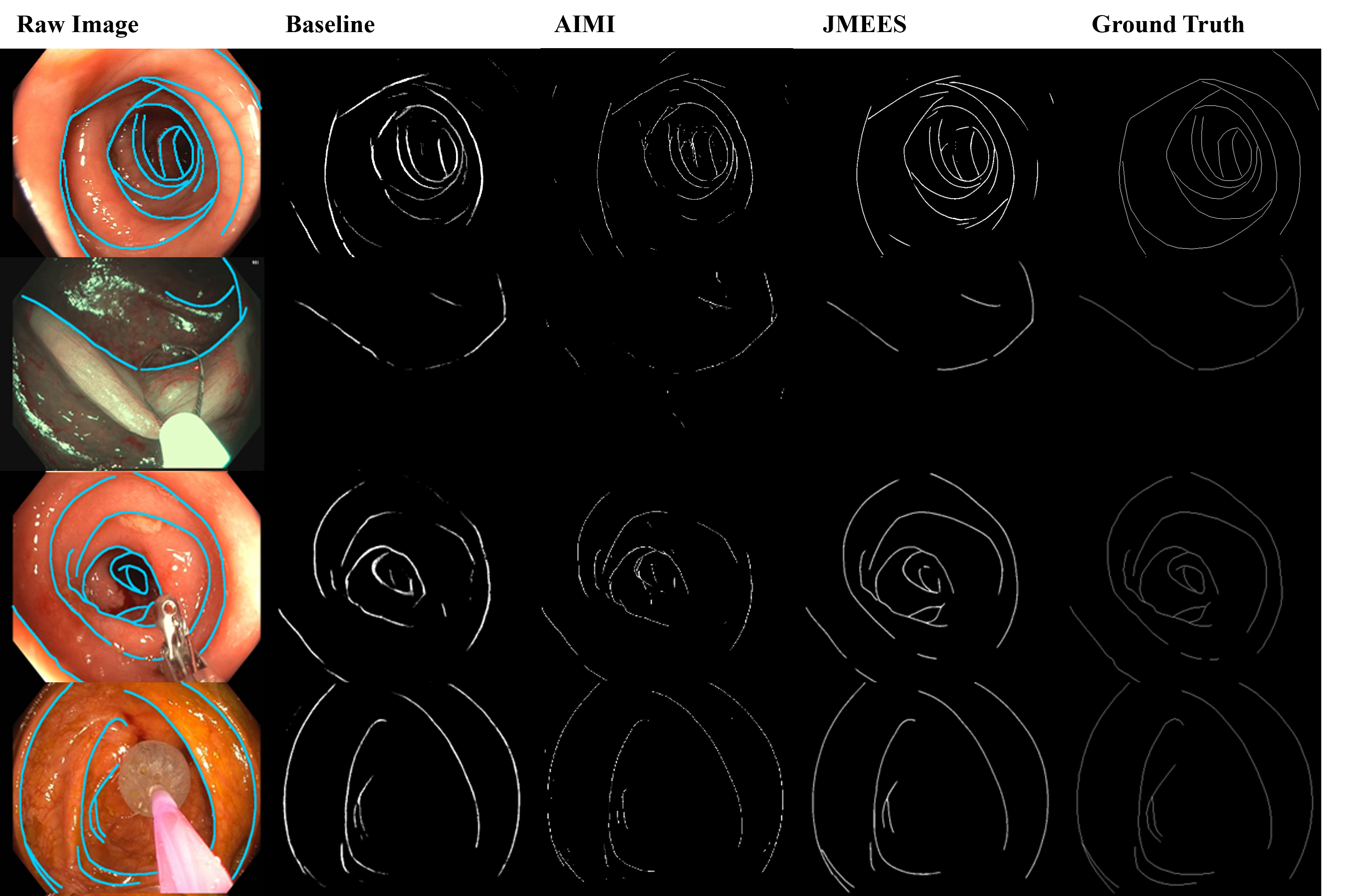}
        \caption{Fold segmentation predictions.}
        \label{fig:segcol_1_1}
    \end{subfigure}\hfill
    \begin{subfigure}[t]{0.51\textwidth}
        \centering
        \includegraphics[width=\linewidth]{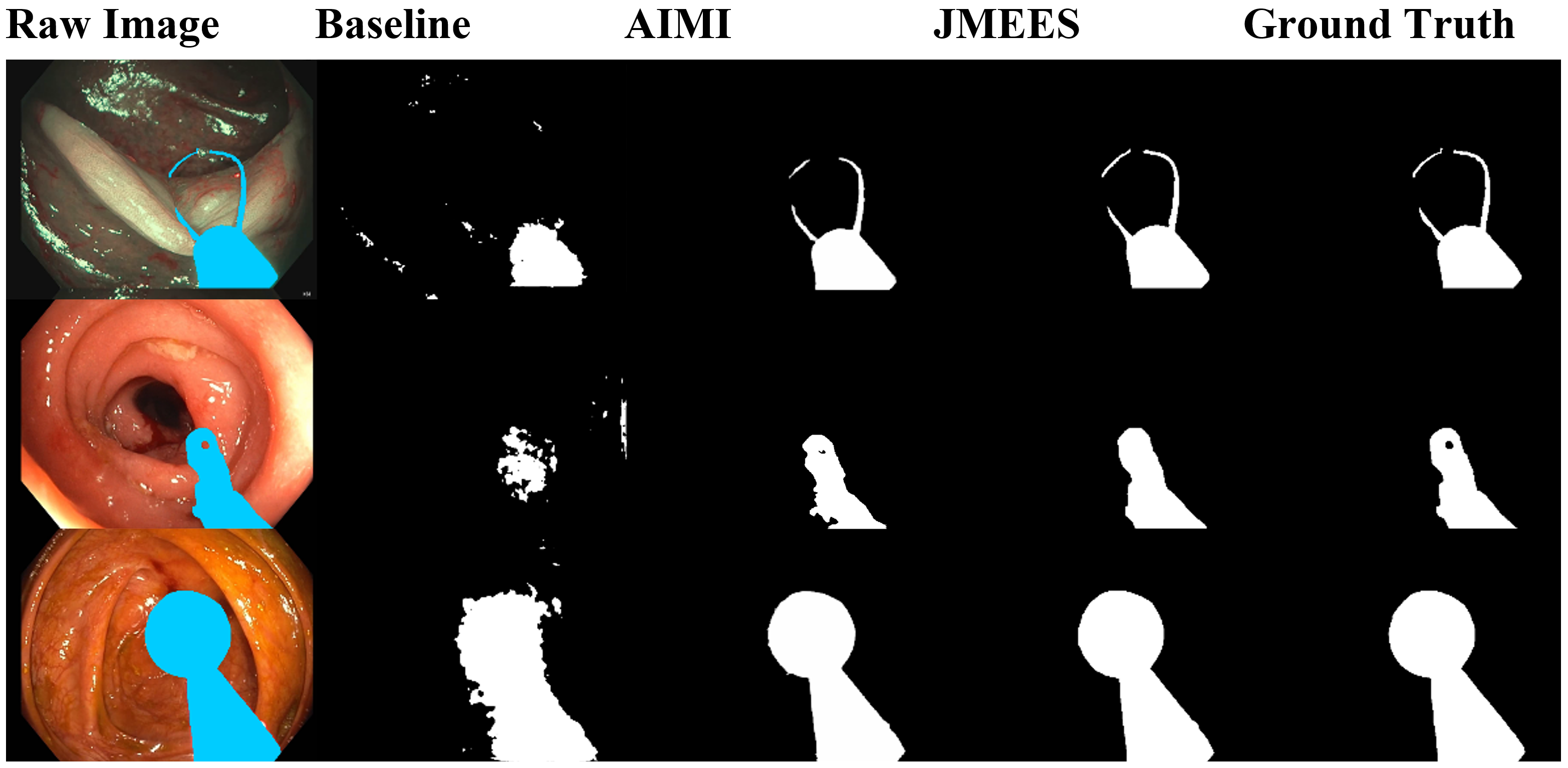}
        \caption{Tool segmentation predictions.}
        \label{fig:segcol_1_2}
    \end{subfigure}

    \caption{
    \textbf{SegCol Task 1 prediction visualizations.}
    Qualitative comparison of fold and tool segmentation results across competing methods. 
    Each row presents the input image, ground truth, and corresponding predictions.
    }
    \label{fig:segcol_task1_vis}
\end{figure}

\textbf{Fold Segmentation:} Jmees achieved the best performance across all five metrics, followed by Dexined (baseline), then AIMI (\Cref{tab:segcol_table1}). These results suggest that fold edge segmentation benefits from architecture designs specifically engineered for this class (Jmees, Dexined) instead of general-purpose semantic segmentation models (AIMI). AIMI's ablation results (\Cref{tab:aimi_ablation}) further reinforce this by showing that regardless of the chosen backbone and architecture, a standard semantic segmentation model has limited performance on the Fold class. Both Jmees and Dexined employ techniques to aggregate predictions at multiple resolutions, albeit in different ways. Dexined combines predictions from downsampled resolutions at different layers, while Jmees utilises a model ensemble using encoders that process either full-resolution images or zoomed-in patches in a sliding window. The zoomed patch aggregation strategy of Jmees proved especially beneficial for handling very thin Fold details that may be lost in Dexined's global downsampling strategy. In addition, Jmees leverages monocular depth estimation as additional input, and it utilises additional supervision via auxiliary tasks: tool presence detection and folds count prediction. Overall, these characteristics make Jmees' approach significantly more complex and computationally heavier than alternatives, but also with a clear advantage in accuracy.

These quantitative findings are further supported by the qualitative results shown in \Cref{fig:segcol_1_1}. Jmees produces the most continuous and well-connected fold edges, with minimal fragmentation and strong structural coherence across long fold trajectories. AIMI generates comparatively thin fold predictions and is able to recover many fine structures, but the outputs are more fragmented and locally discontinuous, likely reflecting the absence of equally strong edge-focused supervision. DexiNed, meanwhile, generally produces sharp contours but with occasional fragmentation.

\textbf{Tool Segmentation:} Jmees achieved the strongest overall performance, with near-perfect Dice (0.981) and AP (0.998), followed by AIMI, while the DeepLab baseline trailed substantially behind (\Cref{tab:segcol_table1}). AIMI, notably, achieves its best performance on Balloon (\Cref{tab:segcol_table1}), a category with comparatively simple and compact geometry. Here, AIMI's emphasis on semantic segmentation plays to its strengths, unlike in fold segmentation. Similar to fold segmentation, tool segmentation benefits from architectures that incorporate sophisticated multi-scale feature extraction. AIMI addresses this with geometry-aware multi-scale modules, rather than relying on generic pyramid pooling approaches such as those used in DeepLab. AIMI's ablation study (\Cref{tab:aimi_ablation}) reinforces this, showing that multiple architecture configurations combining these properties consistently outperform the baseline across tool categories. Jmees further improves on this by combining a transformer-based encoder with global receptive fields and a model ensemble incorporating auxiliary supervision and boundary-aware optimisation, as confirmed by progressive gains in its ablation study (\Cref{tab:jmees_tool_ablation}). This combination is particularly effective for instruments with thin, articulated, or disconnected structures, such as Captivator and Forceps, where Jmees achieves near-perfect scores (\Cref{tab:segcol_table1}).

Qualitatively (\Cref{fig:segcol_1_2}, these trends are clearly reflected in the predicted masks. The DeepLab baseline generally recovers only the coarse extent of the instruments and frequently misses fine structures or produces imprecise boundaries. AIMI substantially improves the delineation of challenging regions, such as the hollow interior of forceps, although occasional local boundary irregularities remain. Jmees produces the cleanest and most structurally coherent masks, preserving thin articulated structures while maintaining smooth contours and consistent object topology, in agreement with its clear quantitative advantage.

\textbf{Overall Findings of Task~1:}
The experimental results reveal a clear distinction between fold and tool segmentation in terms of architectural requirements. Fold segmentation is more sensitive to fine-scale boundary localisation and structural continuity, favouring the boundary- and geometry-aware designs discussed above, as reflected in Jmees' and DexiNed's stronger performance on boundary-sensitive metrics. Tool segmentation, by contrast, is less dependent on boundary precision, with Dice, AP, and CLDice approaching saturation for the strongest methods; here, performance instead tracks structural complexity across instrument categories, with AIMI performing best on categories with simpler, more compact morphology (Balloon) and Jmees on categories with thin, articulated structures (Captivator, Forceps), consistent with the category-level trends discussed above. Together, these results suggest that successful surgical segmentation requires matching model inductive bias to the nature of the target structure: explicit boundary- and geometry-aware modelling for folds, and sensitivity to structural complexity across instrument categories for tools.

Nevertheless, the near-perfect Dice and AP values reported for tool segmentation should be interpreted cautiously. Both metrics are computed through global pixel pooling, which weights frames proportionally to the number of positive pixels. Consequently, frames containing large and clearly visible instruments dominate the final score, while difficult examples involving small, partially occluded, or ambiguous tool appearances contribute relatively little. By contrast, frame-wise metrics such as ODS, OIS, and CLDice evaluate each image independently before aggregation and are therefore more sensitive to local boundary errors and difficult thin-structure cases. 


\begin{figure}[ht]
  \centering
  \includegraphics[width=\linewidth]{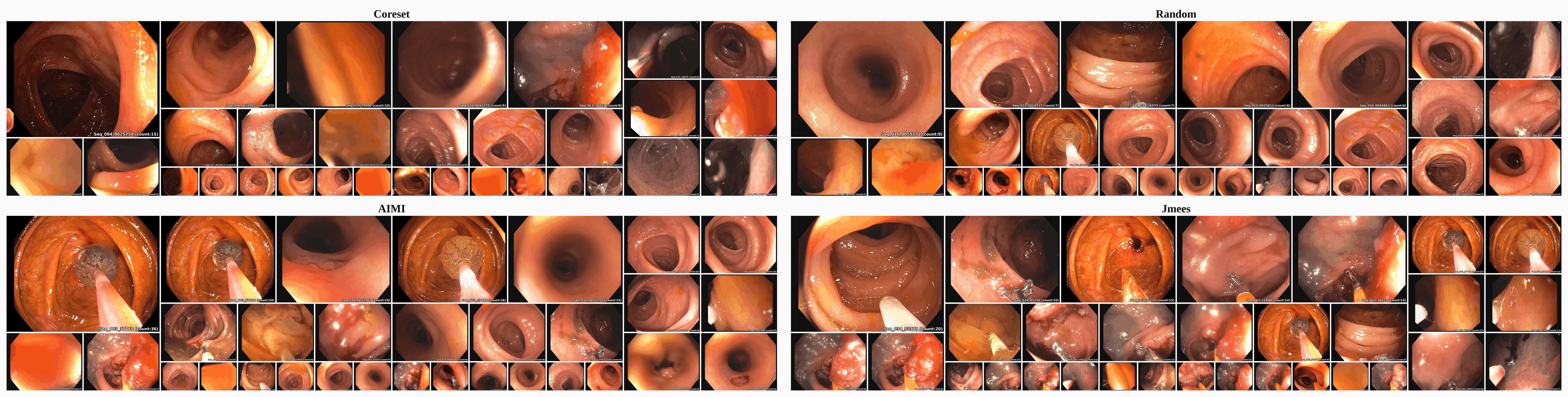}
  \caption{\textbf{Active Learning Sequence Clouds.} Active learning sample images for  participants and baselines (Coreset, Random, AIMI, and Jmees). Each tile represents the most sampled 40-frame clips for each method, and the tile size is proportional to the number of sampled frames for that clip ($n$).}
  \label{fig:al-sequence-cloud}
\end{figure}

\subsection{Task 2}

\textbf{Sequence-level visualization of active learning samples.}
~\Cref{fig:al-sequence-cloud} visualizes the most sampled 40-frame clips. Since each clip will contain many similar images to each other, this figure shows whether algorithms are more greedy (sampling many frames from the most informative clips, even if similar) or attempt to sample for diversity (sampling lower number of frames from many different clips). It also shows which types of images methods prioritise (clear fold views, difficult views, instrument presence, etc).

The visualization reveals clear differences between the sampling strategies. Coreset does not contain a single clip with instruments present in its most sampled sequences. Clips with instruments are both rare and will include very similar frames to each other where the endoscope is kept in place. Therefore any method whose only criteria is sample feature diversity like Coreset will sample few images from clips with instruments. Random sampling only contains a single clip with instruments in its most frequently sampled clips, which just denotes the smaller probability of randomly sampling this under-represented class across the dataset. On the other hand, both AIMI and Jmees include a very high abundance of visible instruments across its most sampled clips, denoting an active targeting of this class.

\begin{table}[ht]
\centering
\caption{SegCol Task2 Quantitative Results: \\ The quantitative results of baseline and participating teams are compared, with the best performances \textbf{bolded} and the second best \underline{underlined}.}

\label{tab:segcol_table2}
\begin{tabular*}{\linewidth}{@{\extracolsep{\fill}} cccccccc}

\midrule
\textbf{Classes} & \textbf{Methods} & \textbf{ODS}$\uparrow$ & \textbf{OIS}$\uparrow$ & \textbf{Dice}$\uparrow$ & \textbf{AP}$\uparrow$ & \textbf{CLDice}$\uparrow$ & \textbf{Threshold} \\
\midrule
\multirow{5}{*}{Folds} & Baseline & 0.250 & 0.258 & 0.265 & \textbf{0.170} & 0.279 & 0.830 \\
 & Random & \underline{0.265} & \textbf{0.275} & \textbf{0.282} & 0.089 & \textbf{0.311} & 0.890 \\
 & Coreset & {0.257} & 0.267 & 0.276 & 0.089 & \underline{0.289} & 0.900 \\
 & AIMI & {0.257} & 0.263 & 0.274 & 0.088 & 0.280 & 0.870 \\
 & Jmees & \textbf{0.267} & \underline{0.273} & \underline{0.281} & \underline{0.092} & \underline{0.289} & 0.870 \\
\midrule
\multirow{5}{*}{Tools} & Baseline
& \textbf{0.235} & \textbf{0.281} & \textbf{0.660} & \textbf{0.699} & \textbf{0.245} & 0.440 \\
 & Random
& 0.181 & 0.215 & 0.496 & 0.260 & 0.160 & 0.620 \\
 & Coreset
& 0.195 & 0.227 & 0.452 & 0.219 & 0.157 & 0.590 \\
 & AIMI
& \underline{0.229} & \underline{0.249} & \underline{0.600} & \underline{0.367} & \underline{0.186} & 0.620 \\
 & Jmees & 0.182 & 0.227 & 0.543 & 0.301 & 0.148 & 0.630 \\
\midrule\midrule
\multirow{5}{*}{Captivator} & Baseline & \textbf{0.077} & \textbf{0.085} & \textbf{0.507} & \textbf{0.430} & \textbf{0.286} & 0.410 \\
 & Random & 0.044 & 0.053 & 0.139 & 0.033 & 0.036 & 0.480 \\
 & Coreset & 0.043 & 0.052 & \underline{0.298} & \underline{0.090} & \underline{0.039} & 0.680 \\
 & AIMI & \underline{0.048} & \underline{0.059} & 0.274 & 0.081 & \underline{0.039} & 0.530 \\
 & Jmees & 0.023 & 0.031 & 0.066 & 0.017 & 0.017 & 0.320 \\
\midrule
\multirow{5}{*}{Forceps} & Baseline & \textbf{0.058} & \textbf{0.082} & \textbf{0.168} & \textbf{0.113} & \textbf{0.253} & 0.440 \\
 & Random & 0.045 & 0.055 & 0.083 & 0.012 & 0.049 & 0.530 \\
 & Coreset & \underline{0.053} & 0.061 & \underline{0.088} & \underline{0.015} & \underline{0.053} & 0.480 \\
 & AIMI & \underline{0.053} & \underline{0.063} & 0.080 & 0.012 & 0.049 & 0.420 \\
 & Jmees & 0.044 & 0.056 & 0.056 & 0.010 & 0.040 & 0.380 \\
\midrule
\multirow{5}{*}{Balloon} & Baseline & \textbf{0.680} & \textbf{0.681} & \textbf{0.829} & \textbf{0.923} & \textbf{0.918} & 0.650 \\
 & Random & 0.208 & 0.212 & \underline{0.707} & \underline{0.503} & 0.343 & 0.690 \\
 & Coreset & 0.118 & 0.125 & 0.477 & 0.262 & 0.248 & 0.570 \\
 & AIMI & \underline{0.282} & \underline{0.284} & 0.678 & 0.479 & \underline{0.413} & 0.650 \\
 & Jmees & 0.172 & 0.173 & 0.678 & 0.478 & 0.317 & 0.630 \\
\midrule
\end{tabular*}
\end{table}

\begin{figure}[ht]
    \centering
    \begin{subfigure}[t]{\textwidth}
        \centering
        \includegraphics[width=0.8\linewidth]{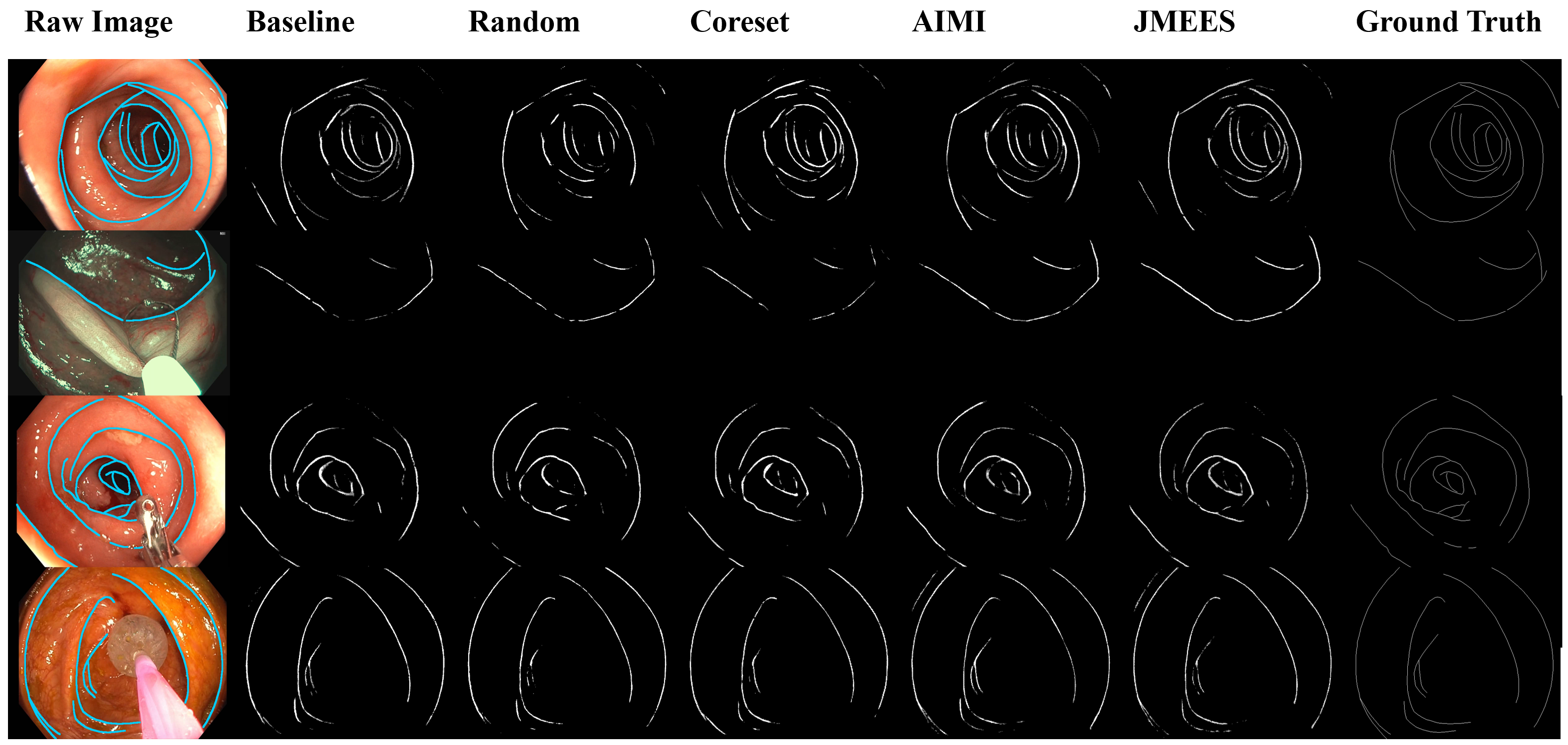}
        \caption{Fold segmentation under active learning strategies.}
        \label{fig:segcol_2_1}
    \end{subfigure}

    \vspace{0.8em}

    \begin{subfigure}[t]{\textwidth}
        \centering
        \includegraphics[width=0.8\linewidth]{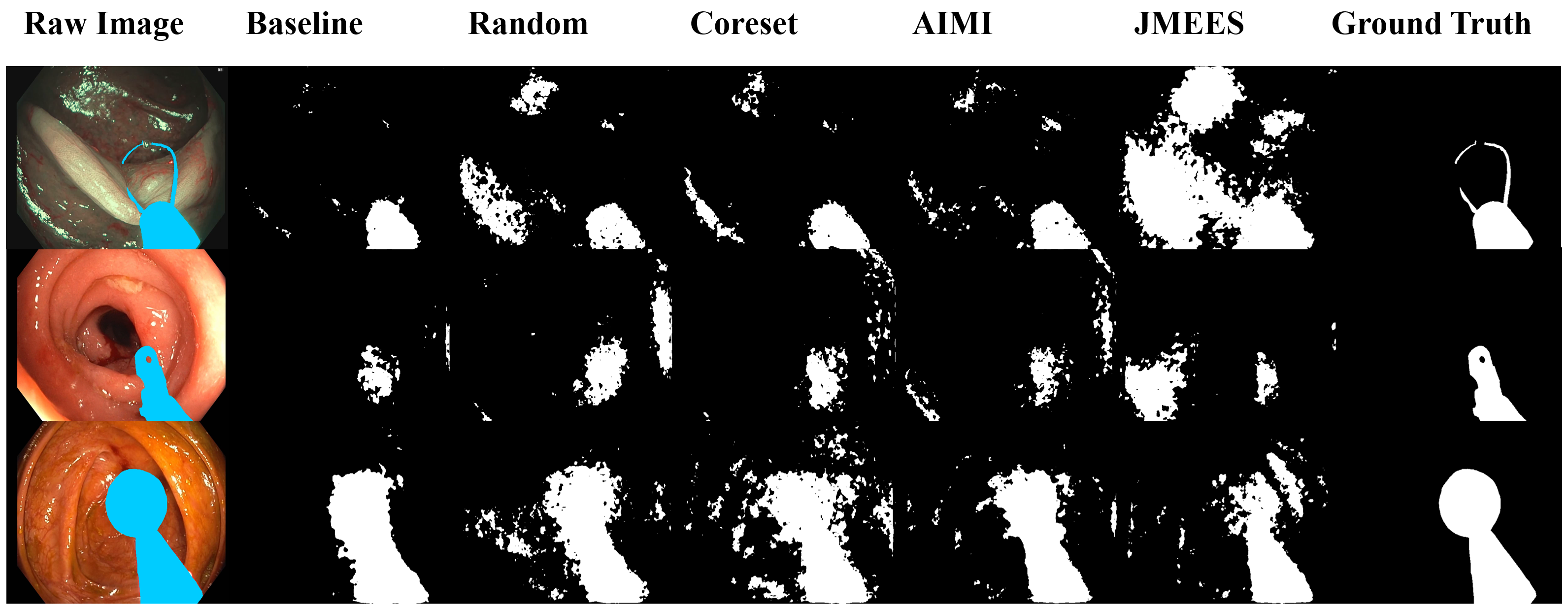}
        \caption{Tool segmentation under active learning strategies.}
        \label{fig:segcol_2_2}
    \end{subfigure}

    \caption{
    \textbf{SegCol Task 2 prediction visualizations.}
    Qualitative comparison of fold and tool segmentation results under different active learning strategies.
    }
    \label{fig:segcol_task2_vis}
\end{figure}

\textbf{Fold Segmentation:}

Adding new training samples has a general positive effect on fold segmentation on most metrics, regardless of active learning strategy. Random sampling and Jmees achieved the strongest performance, which highlights that maximising sample feature variance/entropy (Coreset, AIMI) is not the best active learning criteria in this context. Notably, there are no methods clearly outperforming random sampling. This denotes that, while additional training data is useful for fold segmentation, there is no evidence supporting specific criteria to target specific samples, with classic active learning techniques not being well suited for this application context.

\textbf{Tool Segmentation:}

Active learning methods generally degrade the performance of tool segmentation, regardless of active learning technique. Firstly, Coreset and Random Sampling mostly sample images without tools (Fig. \ref{fig:segcol_task2_vis}), making these classes even more under-represented after active sampling. While AIMI, Jmees frequently sample additional training samples with tools, they still do not sample a balanced distribution of their 3 classes and still lead to worse segmentation models. 

We also note that both the initial training set and the additional data provided are limited to tools from a single endoscopic manufacturer and model, and therefore there is limited added value that can be achieved by increasing tool samples beyond a certain limit, which is likely to have been achieved with the initially curated dataset. This stands in opposition to fold segmentation, where anatomy variability of each patient is extremely large and impossible to fully capture with a small number of frames.

\textbf{Overall Findings of Task~2:}

The overall results suggest that fully automated active learning for colonoscopy scene segmentation is still an open challenge, and that handcrafted processes with a human-in-the-loop such as the one described in Fig. \ref{fig:annotation_pipeline} cannot be easily replaced.

The application of general active learning techniques to tool segmentation is severely limited by the small frequency of these classes within large colonoscopy videos, and the small margin for added value to the initial training dataset released to participants. The results from task 1 also suggest that there is limited margin for tool segmentation improvement. 

A hand-crafted process that directly controls how many samples are included per class, and how many adjacent frames are selected from each sequence, may therefore remain more suitable for addressing class imbalance and temporal redundancy in this setting.

\subsection{Failure Case Analysis}

\begin{figure*}[t]
    \centering

    \begin{subfigure}[t]{0.32\textwidth}
        \centering
        \includegraphics[width=\linewidth]{%
        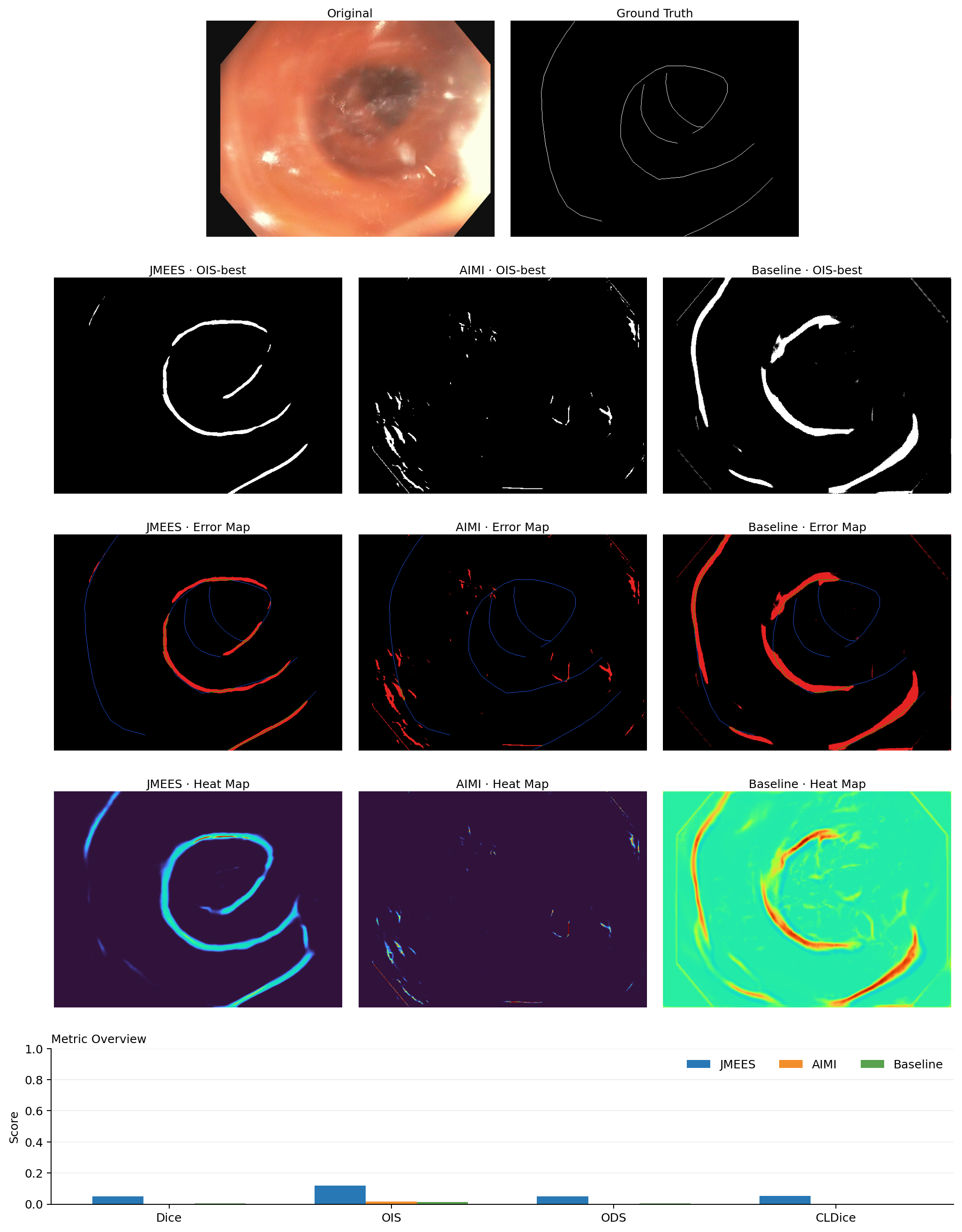}
        \caption{OIS threshold inconsistency.}
        \label{fig:OIS_inconsistency}
    \end{subfigure}
    \hfill
    \begin{subfigure}[t]{0.32\textwidth}
        \centering
        \includegraphics[width=\linewidth]{%
        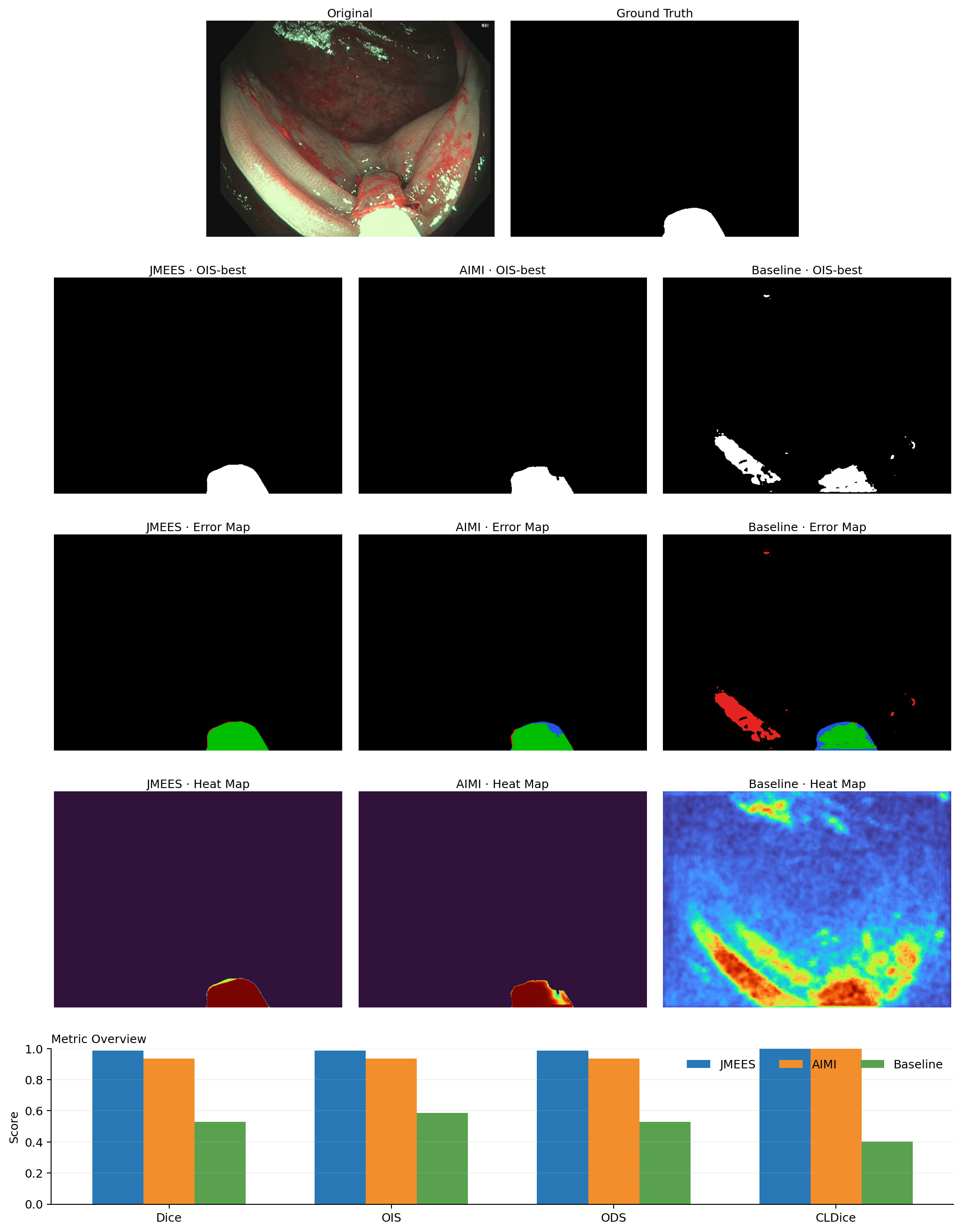}
        \caption{Limited local sensitivity of CLDice.}
        \label{fig:cldice_weakness}
    \end{subfigure}
    \hfill
    \begin{subfigure}[t]{0.32\textwidth}
        \centering
        \includegraphics[width=\linewidth]{%
        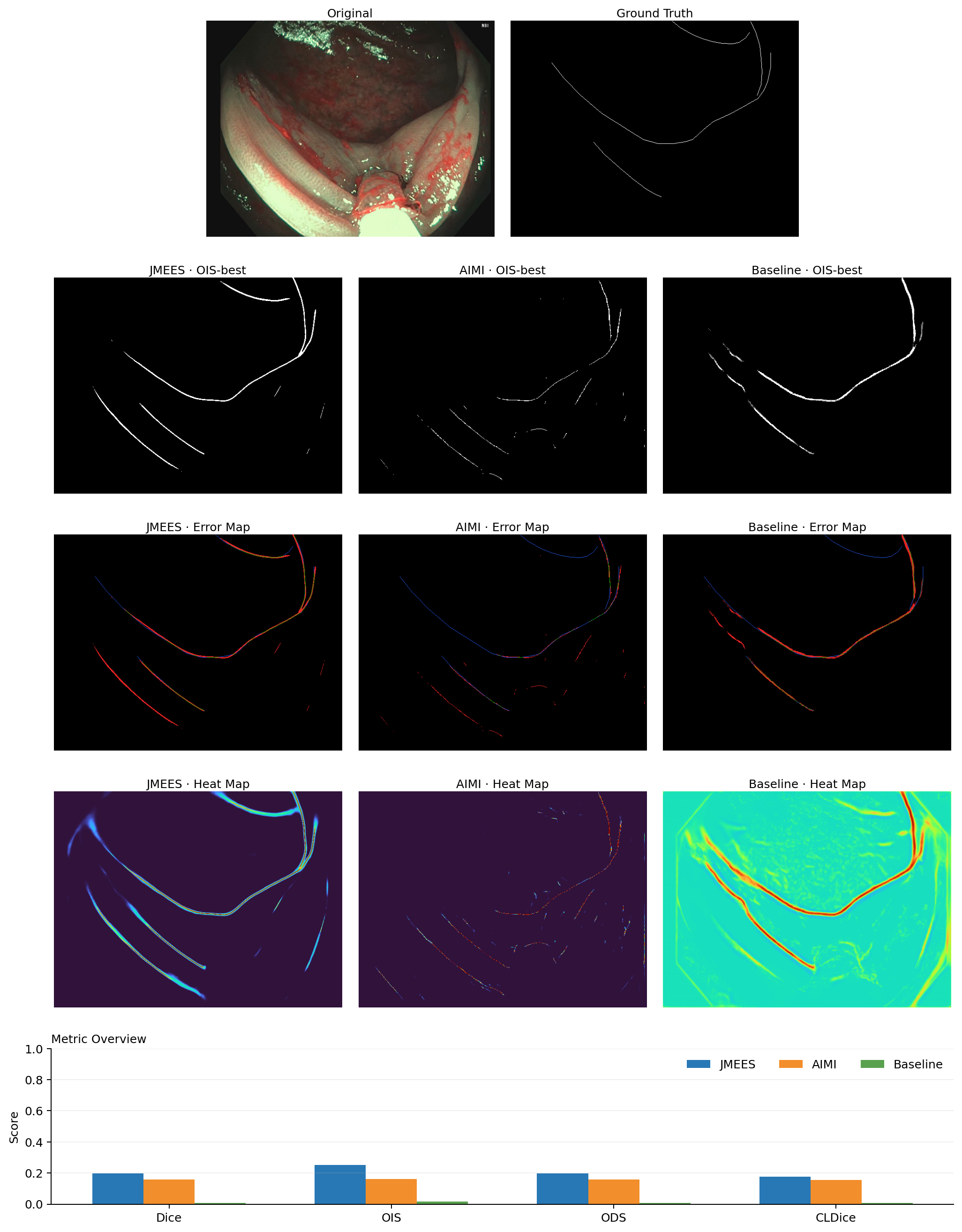}
        \caption{Mismatch between visual quality and metric scores.}
        \label{fig:metrics_limitation}
    \end{subfigure}

    \caption{Representative limitations of segmentation metrics:
    per-image OIS thresholding can reveal positives missed by a universal
    threshold; CLDice can under-reflect subtle local deviations; and visually
    reasonable predictions can receive uniformly low metric scores.}
    \label{fig:metric_limitations}
\end{figure*}

Figure~\ref{fig:OIS_inconsistency} illustrates representative failure patterns in fold segmentation under challenging visual conditions. AIMI shows substantial degradation in blurred regions, where fold responses become fragmented or disappear entirely. This is consistent with the behaviour of DeepPyramid+: although Pyramid View Fusion and Deformable Pyramid Reception improve multi-scale and shape-adaptive representation, they still rely on reliable local boundary evidence. When blur weakens the fold--mucosa transition, the deformable receptive fields may adapt to ambiguous texture rather than true anatomical contours. AIMI is also sensitive to specular highlights, which can either be falsely interpreted as fold-like boundaries or suppress nearby fold predictions.

The baseline method captures visually prominent folds with strong contrast, but misses deeper and thinner folds in defocused regions. This suggests that its predictions are dominated by local edge strength and contrast, with limited ability to recover weak anatomical boundaries once the visual evidence becomes degraded.

Jmees preserves fine-scale fold details more consistently. This advantage is consistent with its RGB--depth input, multi-resolution EfficientNet/EfficientNetV2 branches, patch-based MaxViT inference, and boundary-aware/HED-style supervision, which together provide both geometric cues and explicit contour constraints. However, Jmees also fails near strong highlights in close-up regions, indicating that even depth-augmented and boundary-supervised models remain vulnerable when saturation removes the visual evidence of the fold boundary.

Figure~\ref{fig:cldice_weakness} presents representative failure characteristics in tool segmentation. Jmees achieves near-perfect region overlap, indicating strong localisation from the MaxViT--U-Net++ backbone and Mask2Former ensemble. However, minor boundary errors remain along reflective instrument edges, suggesting that global reasoning and boundary-aware loss improve structural coherence but cannot fully resolve local contour ambiguity caused by specular reflection.

AIMI shows more visible contour fragmentation around highlighted tool regions. Although AdaptNet’s CPF, SSF, and FFD modules support multi-scale extraction, shape/scale-adaptive fusion, and pixel-level feature selection, the method remains sensitive when highlights corrupt the local appearance used for adaptive fusion. The baseline exhibits the most severe failure mode, misclassifying portions of mucosal folds as instruments. This reflects insufficient class-specific shape modelling and an over-reliance on elongated local edge patterns rather than global instrument geometry.

Overall, these examples show that the main failure modes are not merely due to insufficient segmentation capacity. Instead, they arise when photometric artefacts such as blur, specular highlights, and intensity saturation corrupt the local evidence on which both boundary-aware and adaptive-fusion models depend. Architectural advances improve structural completeness under normal conditions, but artefact-aware modelling remains necessary for robust surgical segmentation.

\section{Discussion on Metrics}

The participant results in \Cref{sec:Eval} reveal that different segmentation metrics often favour different methods, structural properties, and failure modes. They are also influenced not only by segmentation quality itself, but also by the underlying evaluation protocol, including aggregation strategy, threshold selection, and structural sensitivity. These observations suggest that commonly used segmentation metrics are not interchangeable and may emphasise fundamentally different aspects of prediction quality, including overlap, contour alignment, confidence calibration, and topological consistency.

Motivated by these inconsistencies, we further investigate the behaviour and limitations of the evaluation metrics used in this challenge. To better understand these effects, we analyse the behaviour and limitations of the metrics used in this challenge from four complementary perspectives: threshold selection, topology-aware evaluation, aggregation strategy, and structural sensitivity. Specifically, we examine how adaptive thresholding affects OIS, how CLDice captures structural connectivity while missing fine local errors, how global pixel pooling and frame-wise averaging influence leaderboard interpretation, and how controlled perturbations reveal metric sensitivity to scale distortion, spatial misalignment, and topology disruption across different object geometries. Our goal is not only to interpret the challenge rankings more carefully, but also to better understand which metrics are most appropriate for evaluating thin anatomical structures and complex surgical instruments.
\subsection{Segmentation Metrics Analysis}

\paragraph{OIS Inconsistency}

Figure~\ref{fig:OIS_inconsistency} shows that the fixed global threshold suppresses low–to–mid confidence responses and yields few/no detected pixels. In contrast, the OIS (per-image optimal) threshold adapts to the image-specific score distribution and recovers plausible positive structures that are visible in the heatmap but remain below the universal threshold. This illustrates why OIS can behave inconsistently across methods and frames: performance may be driven by threshold adaptivity rather than genuinely stronger calibrated confidence, leading to apparent gains (or reversals) that are not reflected under a single operating point.

\paragraph{CLDice Strength and Weakness}

Figure\ref{fig:cldice_weakness} illustrates a characteristic behavior of CLDice in structural evaluation. In the example, the third row contains clear structural/topological errors (e.g., broken or incorrect connectivity), and CLDice penalizes these errors strongly. In contrast, the second row is globally consistent in topology but has subtle local inaccuracies, which are not reflected by the perfect score of CLDice. In short, CLDice is effective for topology-level consistency, but less sensitive to fine-grained detail quality; therefore, it should be interpreted together with region/overlap metrics such as Dice and ODS.

\paragraph{Metrics Limitation} 

In Figure~\ref{fig:metrics_limitation}, even the most structurally complete prediction (Jmees) and the baseline model, which captures the overall geometry but misses portions of the fold region, both obtain consistently low scores across all four metrics. 

In our dataset, the annotated folds have one-pixel width. As a result, small spatial deviations produce disproportionately large false-positive and false-negative regions relative to the narrow foreground area. This leads to low Dice as well as low ODS/OIS scores. Although CLDice is designed to emphasize topological consistency, its computation relies on skeletonization; when the target structure is already close to one-pixel width, the extracted skeleton becomes highly sensitive to even minor boundary misalignment. Consequently, CLDice also remains below 0.2 in this example.

This limitation may cause an underestimation of the practical quality of the predictions without fully reflecting the models’ overall segmentation capability. Nevertheless, it remains discriminative across methods, as relative differences between teams are still preserved under this challenging thin-structure regime.

\paragraph{Global vs Frame-wise Aggregation}
Although Dice and ODS are both overlap, F1-based metrics and often produce nearly identical values at the single-image level, they differ fundamentally in how aggregation is performed across frames. 


Global Dice aggregates true positives, false positives, and false negatives across all pixels in the evaluation set, effectively weighting each frame in proportion to its foreground area. In contrast, ODS is derived from threshold-based boundary evaluation and reports the best single operating point at the dataset level. Its computation depends on boundary matching statistics under a fixed threshold, rather than direct pixel-count aggregation.

As a result, when object size varies substantially across frames, the two metrics can diverge at the sequence or class level. Frames containing thinner or smaller segmentations contribute relatively less to global Dice due to their limited pixel mass, yet small boundary deviations in these frames can significantly affect ODS. Consequently, performance degradation on thin structures tends to induce a stronger penalty in ODS than in global Dice, even when per-image values appear numerically similar.

This distinction explains why Dice and ODS may appear numerically similar for individual images yet diverge at the sequence or class level when object sizes vary substantially across frames. Because thinner structures are more sensitive to error, degradation in such frames has a disproportionately larger impact on ODS compared to global Dice. As a result, ODS tends to exhibit stronger penalization under thin structures. This explains the difference between Dice and ODS results in \Cref{tab:segcol_table1}.

\begin{figure}[ht]
    \centering
        \centering
        \includegraphics[width=\linewidth]{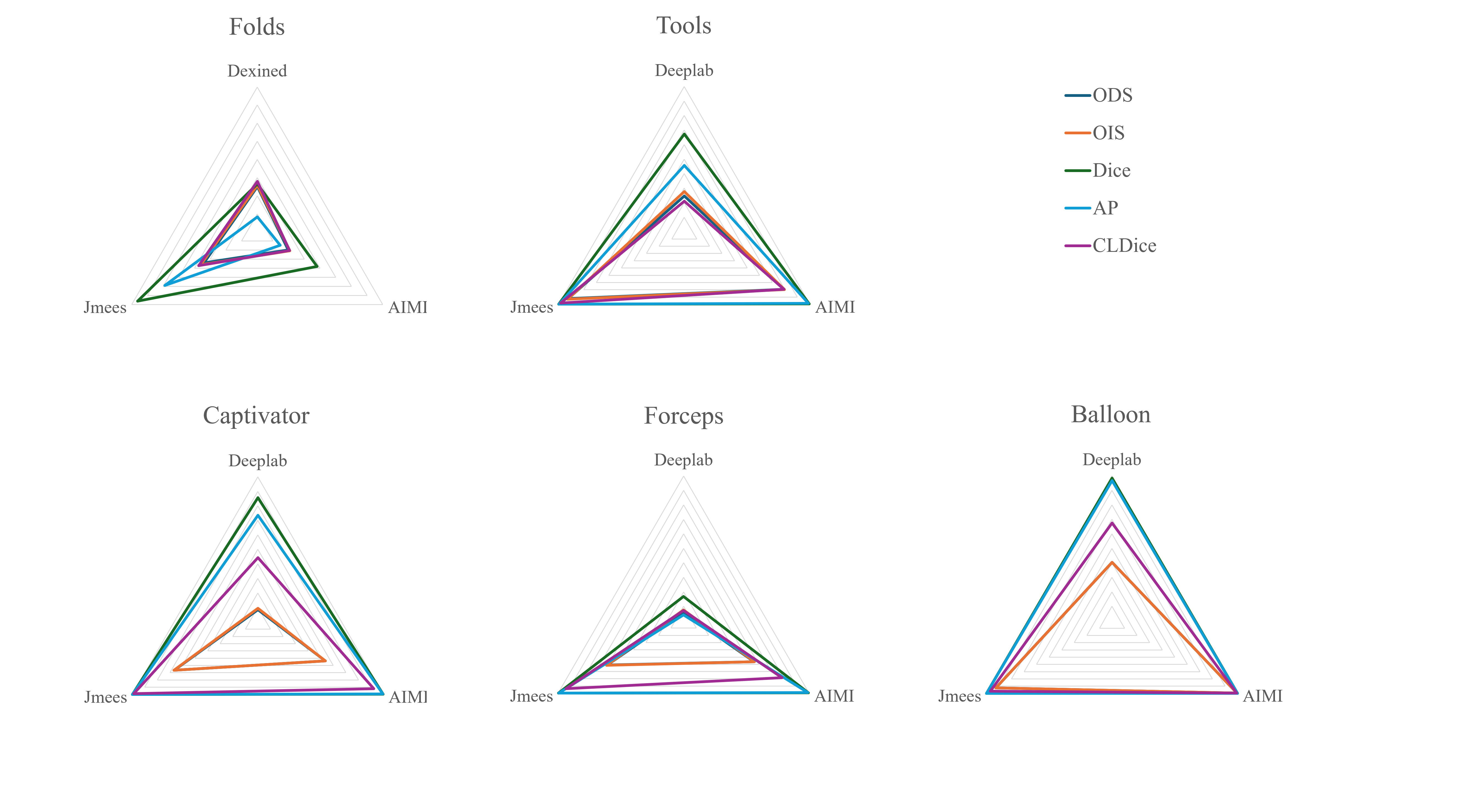}

    \caption{
    \textbf{Visual comparison of segmentation (Task 1)} 
    Radar plots summarize results across five metrics (ODS, OIS, Dice, AP, CLDice), where each vertex denotes a competing team or strategy and larger radial values indicate stronger performance. 
    }
    \label{fig:task1_task2_vis}
\end{figure}

From this pattern, the metrics can be grouped into two categories: global metrics (AP and Dice) and frame-based metrics (OIS, ODS, and CLDice). Global metrics aggregate information across the dataset, whereas frame-based metrics evaluate performance at the image level before averaging. As shown in Task 1 (\Cref{fig:task1_task2_vis}), the global metrics appear more tolerant to local segmentation errors and exhibit larger performance improvements compared to the frame-based metrics.

\subsection{Systematic Perturbation Analysis}
\paragraph{Perturbation Design and Evaluation Protocol}
\begin{figure}
    \centering
    \includegraphics[width=\linewidth]{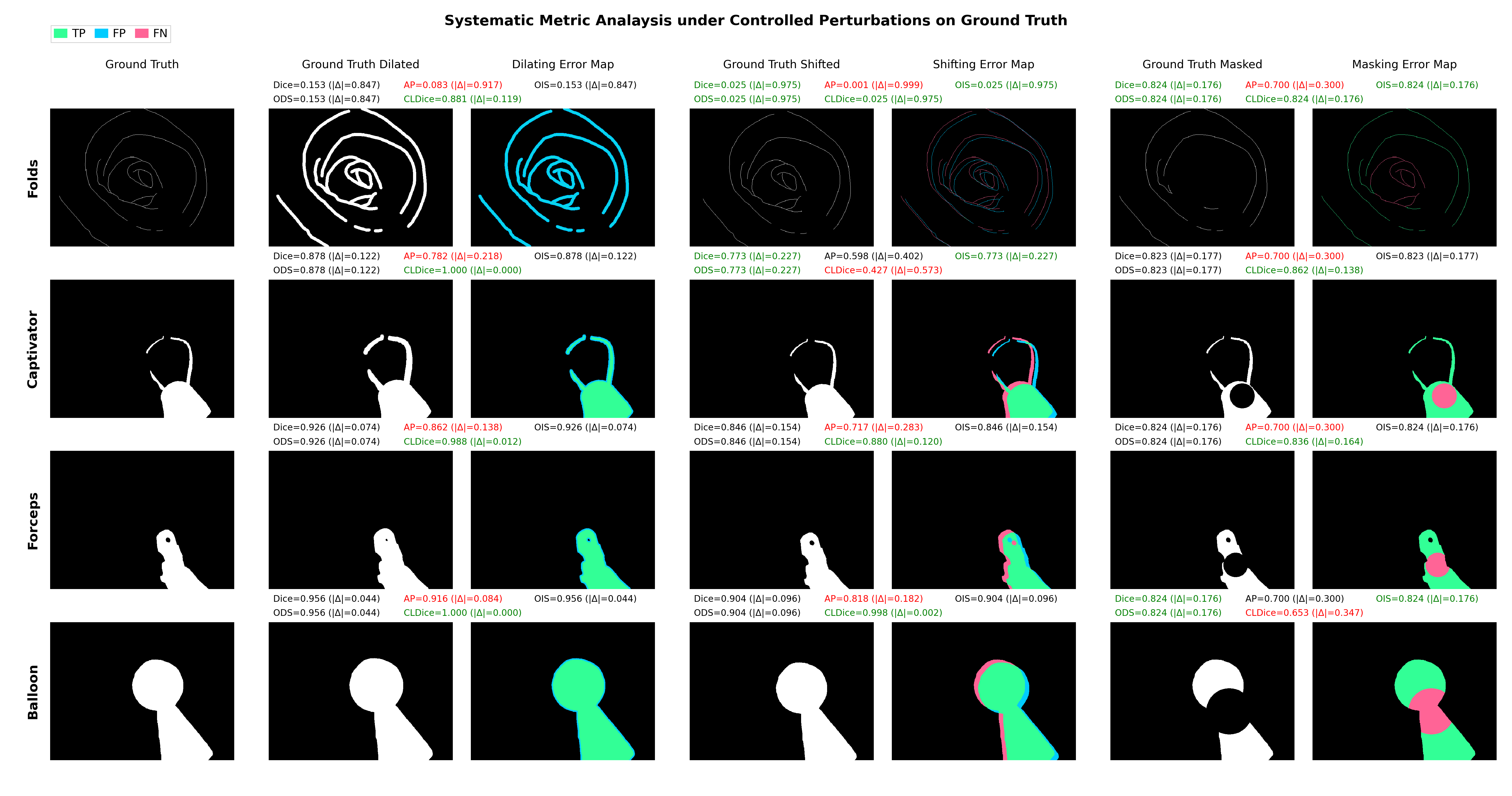}
    \caption{\textbf{Systematic metric analysis under controlled perturbations applied to ground-truth masks.} For each anatomical or tool category (rows), three types of structured perturbations are introduced to the ground truth: dilation, spatial shift, and localized inpainting (single-disk occlusion). For each operation, the perturbed mask and corresponding error map (TP: green, FP: cyan, FN: pink) are shown. Per-image metrics (Dice, CLDice, OIS, ODS) are reported relative to the original ground truth baseline, with color coding indicating relative deviation magnitude, red indicating changing the most while green the least.}
    \label{fig:systemtic_analysis}
\end{figure}

\begin{figure}
    \centering
    \includegraphics[width=\linewidth]{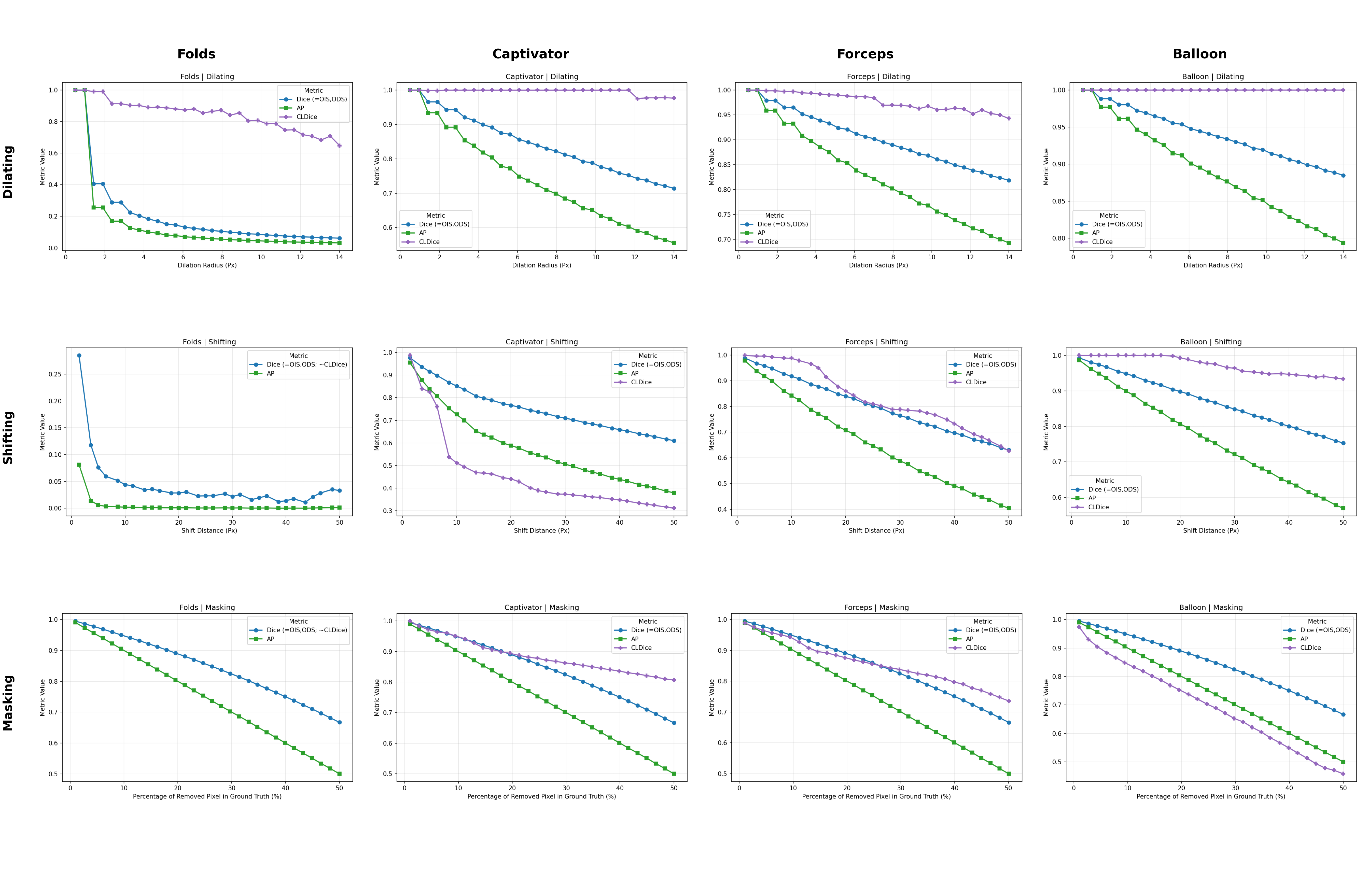}
    \caption{\textbf{Systematic perturbation analysis of 5 segmentation metrics across 4 instrument classes under 3 controlled ground-truth operations.} Metric responses are shown across increasing perturbation severity. Curves whose values are numerically identical across all perturbation levels (difference $<10^{-10}$, denoted by `=') or nearly identical (maximum absolute difference $<5\times10^{-4}$, denoted by `$\sim$') are merged and indicated accordingly in the legend.}
    \label{fig:systemtic_analysis_chart}
\end{figure}

To better understand the structural sensitivity of commonly used segmentation metrics, we systematically perturb the ground-truth masks along three complementary deformation axes. Specifically, we simulate:
\begin{itemize}
    \item Area expansion via morphological \textit{dilation}, which enlarges the predicted region while largely preserving its overall shape;
    \item Spatial displacement through controlled mask \textit{shifting}, which alters object localization without modifying its internal structure; and
    \item Localized occlusion by removing a single contiguous region corresponding to a fixed proportion of object area (\textit{masking}), thereby introducing structured annotation degradation.
\end{itemize}

Although these perturbations are synthetic, they approximate common annotation and model failure modes: boundary over-segmentation (dilation), localization drift (shifting), and partial occlusion or missing annotation (masking).

These controlled perturbations are designed to disentangle how each metric responds to distinct types of structural deviations, where dilation corresponds to scale distortion, shifting corresponds to spatial misalignment, and masking corresponds to topology disruption through removal of connected regions.

Figure~\ref{fig:systemtic_analysis} visualizes representative examples under the three perturbation types. In each sub-panel, we annotate Dice, ODS, OIS, CLDice, and AP. We color the metric with the largest $\Delta$Metric in red and the smallest $\Delta$Metric in green. We use black when the metrics show the same $\Delta$Metric.

Figure~\ref{fig:systemtic_analysis_chart} evaluates metric sensitivity under structured annotation perturbations by varying perturbation intensity in 30 discrete steps for each operation and computing per-class metric trends on representative single-image cases.

\paragraph{Structural Regimes: Thickness and Topological Complexity}

We analyze four categories in Task 1 that span distinct geometric and topological regimes. Importantly, the variation is not solely along a thickness axis, but also along a structural complexity axis.

\begin{itemize}
    \item \textbf{Folds}: extremely thin, elongated structures with multiple connected branches and high boundary-to-area ratio.
    \item \textbf{Captivator}: composite structures containing both thin and thick components.
    \item \textbf{Forceps}: larger regions with holes and moderate topological complexity.
    \item \textbf{Balloon}: predominantly thick, compact, single-component regions with simple topology.
\end{itemize}

This ordering establishes a controlled geometry axis from thin–highly-branched to thick–topologically-simple objects, enabling us to analyze how both thickness and connectivity redundancy influence metric behaviour.

\textbf{Metric Response by Perturbation} %

\textit{Scale Distortion (Dilation):} Under dilation, all overlap-based metrics exhibit progressively reduced sensitivity as structural thickness increases. Thin geometries, such as folds, undergo substantial structural alteration even under small dilation radii, whereas thick compact objects retain most of their intersection area. AP consistently exhibits the steepest degradation under dilation across all categories, indicating the highest sensitivity to scale expansion. CLDice exhibits minimal variation across dilation levels. This behaviour can be explained by its reliance on skeleton overlap: CLDice depends on centerline consistency, and uniform dilation does not alter the underlying skeleton topology. Consequently, skeleton overlap remains largely preserved even when volumetric overlap increases artificially.

\textit{Spatial Misalignment (Shifting):} Under spatial shifting, extremely thin structures demonstrate abrupt metric decay even under small offsets, due to rapid loss of intersection. As objects become thicker or contain redundant branches, partial overlap is retained despite displacement, leading to more gradual degradation. Across all categories, AP again shows the strongest penalization under shifting, confirming its overall highest sensitivity among evaluated metrics. CLDice demonstrates relative robustness when connectivity remains preserved. Rigid translation does not break topological connectivity; therefore, as long as partial skeleton overlap remains, CLDice degrades more slowly compared to purely overlap-based metrics.

\textit{Topology Disruption (Masking):} Masking produces a complementary pattern. Thinner multi-branch structures appear more tolerant to localized occlusion, since removal of a single region often affects only one branch while preserving global connectivity. In contrast, thick compact objects such as Balloon experience stronger degradation, as removal of a contiguous region may sever the primary structural backbone. AP again demonstrates the largest magnitude of degradation across masking intensities, reinforcing its global sensitivity across perturbation types. CLDice exhibits asymmetric behaviour:

When masking disrupts the principal centerline in topologically simple objects, CLDice drops substantially. However, in thin, highly branched structures with redundant connectivity, the global skeleton can remain intact despite local removal, resulting in comparatively smaller degradation.

\paragraph{Cross-Metric Interpretation and Implications}
Overall, our results show that metric sensitivity is jointly determined by object geometry and perturbation type, and therefore cannot be interpreted independently of structural context. Overlap-based metrics such as AP, Dice, ODS, and OIS primarily reflect retained spatial intersection and thus favor thicker structures under dilation and shifting, where intersection decays more gradually as width increases. In contrast, thin structures with high boundary-to-area ratios experience rapid overlap loss under even minor expansion or displacement. Under localized occlusion, however, these same overlap-based metrics often appear more tolerant for thin, highly branched geometries, since removal of a single region affects a smaller proportion of total area. 

Among all evaluated metrics, AP consistently exhibits the greatest sensitivity under all perturbation types and across all structural regimes, making it the most conservative metric in detecting structural deviation. CLDice exhibits a complementary and topology-aware behaviour: 1) it is comparatively insensitive to moderate dilation, as centerline topology is often preserved; 2) it is relatively robust to moderate shifting when connectivity is preserved; 3) it becomes highly sensitive under masking when centerline continuity is disrupted.

Consequently, metric rankings are not invariant to the dominant failure mode of a segmentation model: a method that appears robust under scale distortion or misalignment may be unstable under topology disruption, and vice versa. These findings underscore that no single metric provides a geometry-agnostic assessment of segmentation quality; instead, metric selection should be explicitly aligned with whether spatial overlap or structural connectivity is the principal objective of the task.

\subsection{Overall Findings}
Overall, the metric analysis shows that segmentation performance may appear very different depending on what type of error the metric emphasizes and what structure is being evaluated. Metrics that appear numerically similar under standard cases can respond very differently once predictions exhibit boundary inflation, spatial misalignment, or topology disruption.

A key observation is that thin anatomical structures fundamentally amplify metric sensitivity. For one-pixel-wide folds, even small spatial deviations can cause severe penalties across overlap- and boundary-based metrics despite preserving the visually correct trajectory. In contrast, thicker compact structures remain comparatively stable under similar perturbations because most foreground overlap is retained. Consequently, the same segmentation error may produce disproportionately different scores depending on object geometry.

The analysis also reveals that pooled metrics such as Dice and AP can systematically overestimate performance on dominant or large foreground regions, whereas frame-wise and boundary-sensitive metrics more effectively expose local failure cases and thin-structure instability. Similarly, CLDice captures structural continuity but may remain insensitive to subtle local contour inaccuracies when global connectivity is preserved.

Taken together, these findings suggest that reliable surgical segmentation evaluation requires combining multiple metric families rather than relying on a single score. Region- and confidence-based metrics (Dice and AP) are useful for measuring coarse agreement and dominant-region accuracy, boundary-sensitive metrics (ODS and OIS) better capture contour precision and frame-level robustness, and topology-aware metrics (CLDice) better reflect structural continuity. In particular, evaluation of thin anatomical structures should not rely solely on Dice/AP, since these metrics may fail to capture clinically relevant boundary misalignment and connectivity errors. Reliable interpretation therefore requires matching metric choice to both target geometry and the dominant failure mode.

\section{Conclusion}

In this paper, we introduced the SegCol dataset and challenge for simultaneous fold and surgical instrument segmentation, together with a unified evaluation pipeline spanning supervised segmentation, active learning, and metric sensitivity analysis.

The challenge results show that segmentation performance depends strongly on the interaction between target geometry and model inductive bias. For fold segmentation, the strongest methods consistently rely on explicit boundary- and geometry-aware designs. Jmees achieves the best overall performance by combining RGB--depth fusion, multi-resolution U-Net++ ensembles, patch-based MaxViT inference, and boundary-aware supervision, which together improve the preservation of thin and elongated fold trajectories. DexiNed remains competitive under boundary-sensitive metrics because its edge-oriented formulation directly favours contour localisation, whereas AIMI’s deformable pyramid fusion improves coarse structural coverage but is less effective at maintaining long-range fold continuity under blur and low contrast.

Tool segmentation exhibits a different regime. Compact and visually distinctive instruments such as Balloon are segmented reliably by most methods, with AIMI achieving the strongest results through shape- and scale-adaptive fusion in AdaptNet. In contrast, articulated tools such as Captivator and Forceps benefit more from architectures with stronger global reasoning and feature refinement. Jmees therefore performs best on these categories through the combination of MaxViT-based contextual modelling, U-Net++ refinement, auxiliary classification supervision, and Mask2Former ensemble reasoning. These findings suggest that thin and topologically complex structures benefit more from explicit contour preservation and global contextual reasoning, whereas thick compact structures are comparatively tolerant to local boundary variation.

The active learning benchmark further demonstrates that sample acquisition strategy interacts strongly with dataset redundancy and target structure. Random sampling performs competitively for fold segmentation because neighbouring endoscopic frames contain substantial visual and anatomical overlap. AIMI’s entropy-based strategy preferentially selects uncertain balloon-dominant frames, whereas Jmees’ PCA--kNN--Leiden pipeline increases appearance diversity but often selects wall-dominant views that provide limited additional segmentation difficulty. Consequently, neither uncertainty-based nor diversity-based acquisition consistently improves over random sampling. For tool segmentation, all active learning strategies slightly degrade performance relative to the baseline, suggesting that the selected samples fail to provide sufficient new instrument appearance information and may exacerbate class imbalance under limited annotation budgets.

Beyond leaderboard comparison, the metric analysis reveals that segmentation scores are highly dependent on both object geometry and evaluation protocol. Dice and AP mainly reflect pooled overlap and confidence ranking behaviour, making them comparatively tolerant to local errors in large foreground regions. ODS and OIS are more sensitive to thresholded boundary behaviour and thin-structure localization, while CLDice primarily captures skeleton-level connectivity and is less sensitive to moderate boundary inflation. Controlled perturbation experiments further show that thin structures are disproportionately penalized under dilation and spatial shifting, whereas topology-aware metrics become most sensitive when centerline continuity is disrupted. As a result, different metrics may favour fundamentally different structural properties and failure modes, indicating that leaderboard rankings are not geometry-agnostic.

Overall, SegCol highlights that robust surgical segmentation requires jointly considering target geometry, model design, sampling strategy, and evaluation protocol. Future work may explore geometry-aware supervision, artefact-aware learning, and task-specific active learning strategies that explicitly balance uncertainty, structural diversity, and class coverage. Equally importantly, future surgical benchmarks should adopt evaluation protocols that better capture clinically relevant failure modes for thin and topologically complex structures.


\bibliographystyle{unsrtnat}
\bibliography{ref}

\end{document}